\documentclass[preprint,12pt,3p]{elsarticle}

\usepackage{amssymb}

\usepackage{multirow}
\usepackage{graphicx}
\usepackage[table]{xcolor}
\usepackage{adjustbox}
\usepackage{setspace}
\usepackage{todonotes}
\usepackage{caption}
\usepackage{subcaption}

\journal{Image and Vision Computing -- Biometrics in the Wild}

\begin{document}

\begin{frontmatter}

\title{Found a good match: should I keep searching? -- Accuracy and Performance in Iris Matching Using 1-to-First Search}

\author[label1]{Andrey Kuehlkamp\corref{cor1}}
\address[label1]{University of Notre Dame - Notre Dame, IN}

\cortext[cor1]{Corresponding author}

\ead{akuehlka@nd.edu}

\author[label1]{Kevin Bowyer}
\ead{kwb@nd.edu}

\begin{abstract}
Iris recognition is used in many applications around the world, with enrollment sizes as large as over one billion persons in India's Aadhaar program. 
Large enrollment sizes can require special optimizations in order to achieve fast database searches. 
One such optimization that has been used in some operational scenarios is 1:First search.  
In this approach, instead of scanning the entire database, the search is terminated when the first sufficiently good match is found. This saves time, but ignores potentially better matches that may exist in the unexamined portion of the enrollments. At least one prominent and successful border-crossing program used this approach for nearly a decade, in order to allow users a fast ``token-free'' search.  Our work investigates the search accuracy of 1:First and compares it to the traditional 1:N search.  Several different scenarios are considered trying to emulate real environments as best as possible: a range of enrollment sizes, closed- and open-set configurations, two iris matchers, and different permutations of the galleries. Results confirm the expected accuracy degradation using 1:First search, and also allow us to identify acceptable working parameters where significant search time reduction is achieved, while maintaining accuracy similar to 1:N search.
\end{abstract}


\end{frontmatter}


\section{Introduction}
\label{sec:intro}


One of the most powerful biometric modes  is \textit{Iris Recognition}. 
It is based on images from the area of the eye surrounding the pupil, called the iris. 
Each iris contains a complex pattern composed of elements like crypts, freckles, filaments, furrows, pits, striations and rings. These texture details are what make the iris particularly useful for recognition \cite{Jain2011}.

Since its first demonstration by Daugman \cite{Daugman2008}, iris recognition has evolved to become one of the best-known biometric characteristics. 
The largest biometric database in the world, the Aadhaar program in India, has already collected 1.13 billion people's irises (and fingerprints) for enrollment \cite{latimes2017}. 

In 2016, Somaliland started to register voters using iris biometrics \citep{interpeace2016somaliland}. 
The motivation is to prevent voting fraud, after authorities found a large number of duplicate registrations, even with the use of facial and fingerprint recognition \citep{newsci2014somaliland}.
The decision was made after months of testing and preparation, aided by a feasibility study by a team of academic researchers \citep{bowyer2015somaliland}.

Since 2002, countries like the United Kingdom, Canada and Singapore have used iris biometric systems to perform border-crossing checks on frequent travelers. Similarly, the United Arab Emirates (UAE) has employed an iris-based biometric system to keep track of banned travelers since 2001. 
The UAE system is known for performing approximately 14 billion IrisCode comparisons daily \citep{Daugman2015airports}.


Iris recognition is being deployed in an increasing number of applications, and with larger and larger database sizes.
Although iris matching can be performed in an extremely rapid manner, the need for optimization becomes stronger as the number of enrolled persons in applications becomes larger.
In this sense, we analyze one search technique that is known to have been utilized in some operational scenarios, but whose performance and accuracy have not been considered in the research literature.

In iris databases, the traditional search approach for identification is called 1:N, which means the entire biometric enrollment is scanned and the best match is selected.
For nearly a decade, the NEXUS border-crossing program \citep{nexus1} employed a variation of this search technique, called 1:First, in order to improve search speed.
In 1:First search, the search of the biometric enrollment is terminated when the first biometric template that satisfies the matching threshold is found.
This approach generally speeds up the search.
However, the biometric template selected by this approach may not be the best match.
The biometric template matched in 1:First search is more likely to not correspond to the same subject as does the biometric probe than in 1:N search.

\citet{kuehlkamp2016_1first} found a significant difference in the 1:First False Match Rate (FMR) in comparison to 1:N search, especially with larger enrollment sizes, and higher rotation tolerances.
However, as pointed out by the authors, \citep{kuehlkamp2016_1first} was not complete in some aspects.
The enrollment sizes used in the experiments were rather small.
The evaluation was not done using a ``commercial quality'' iris matcher.
And the experiments did not contemplate open-set scenarios.
The objective of this work is to address each of these issues. 

In this work, experiments are performed on two iris matchers, the second being a well-known commercial matcher.
Also, we use data augmentation techniques to increase the size of our dataset, and perform experiments in enrollment sizes that can be more representative of real world applications.
This also allows us to create and test open-set scenarios. 
In addition, we perform experiments using different permutations of the same enrollment, in order to verify how that could affect 1:First accuracy.


\section{Background and Related Works}
\label{sec:related}


Iris recognition, like other biometric modalities, can be used in two types of identity management functionality \citep{Jain2011}: verification and identification.

In \textbf{verification}, the task of the system is to verify if the identity claimed by the user is true. 
In this case, the biometric reference from the user is compared to a single biometric probe in the database (\textit{one-to-one matching} \citep{Biometrics2006}). 
In turn, when performing \textbf{identification}, the user does not claim an identity. 
Consequently, the system has to compare the user's biometric probe with the biometric references of potentially all the persons enrolled in the database (\textit{one-to-many matching}\citep{Biometrics2006}).

Within identification, it is possible to distinguish \emph{closed-set} and \emph{open-set} identification tasks. 
With a closed-set, the user is known to be enrolled in the database, and the system is responsible to determine his or her identity. 
On the other hand, when doing open-set identification, the system must, before trying to identify a user, determine if he or she is enrolled in the database \citep{Biometrics2006}. 
This work is concerned with one-to-many matching as used in an identification system, and particularly, with exploring the difference between two possible implementations of one-to-many.

The \emph{comparison} procedure is a core part of every biometric identification or verification system. 
In this procedure, the system compares the biometric probes acquired from the user against previously stored biometric references and scores the level of similarity or dissimilarity between them. 
According to a predetermined threshold, the system then makes a decision about the user: either it is a \emph{match} or a \emph{non-match}. 
Declaring a match means to assume that the system accepts both biometric samples as being originated by the same human source \citep{Biometrics2006}.

\subsection{Iris Comparison Output}
\label{sec:matchingoutput}

Two types of errors can be made by biometric systems: \textit{False Match} (FM) and \textit{False Non-Match} (FNM).
A FM occurs when biometric probe and biometric reference from different individuals are incorrectly classified as a match. 
Conversely, FNM occurs when biometric probe and biometric reference of the same individual are not recognized as a match  \cite{Jain2011}.

These errors are very similar to \textit{Type I} (false-positive) and \textit{Type II} (false-negative) statistical errors. 
However, this traditional standpoint usually does not contemplate a scenario variation: open-set \textit{vs.} closed-set \cite{ISO19795}.
In both of these cases, there is an enrollment $G$ of biometric references, and the comparisons made against that enrollment come from the biometric probe set $P$.
If the identities in $P$ are a proper subset of $G$ ($P \subseteq G$), then the scenario is said to be \textit{closed-set}. 
On the other hand, if any of the identities in $P$ are not contained in $G$, that is, $P = \{P \cap G \wedge P \not\subseteq G\}$, the scenario is called \textit{open-set}.

The different search methods, 1:N and 1:First, can produce different results for the same biometric probe and list of biometric references \citep{kuehlkamp2016_1first}.
These situations are described in detail in \citep{kuehlkamp2016_1first}.
Ultimately, this distinction is the source of the accuracy difference between 1:N and 1:First.

\subsection{Comparison Output in Closed and Open Set Scenarios}
\label{sec:matchoutscenarios}

As mentioned by \citet{ISO19795}, and shown in Table \ref{tab:rmatchout}, the conventional definition of comparison results is a little different when considering open-set and closed-set scenarios.
In a closed-set scenario, we have the typical cases of TM and FNM for biometric references. 
An interesting peculiarity of the closed-set scenario is that TNMs cannot happen, because all the references are enrolled (Tab. \ref{tab:rmatchout}$^b$).

On the other hand, in an open-set scenario, all four typical cases occur, but there is a distinction to be made: false matches (Tab. \ref{tab:rmatchout}$^c$) can occur either as Enrolled False Matches (EFMs), like in a closed-set, or as Unenrolled False Matches (UFM), when a biometric probe of an unenrolled individual is similar enough to match one of the enrolled biometric references.

\begin{table}[h]
    \caption{Possible outputs for matching against an enrollment in Closed-set and Open-set scenarios.}
    \label{tab:rmatchout}
    \centering
    \resizebox{\linewidth}{!}{
        \begin{tabular}{ l | c | c | c | c } 
        \hline
         & \multicolumn{4}{|c}{Matching Result} \\ \cline{2-5}
         & \multicolumn{2}{|c}{ Closed-set } & \multicolumn{2}{|c}{ Open-set } \\ \cline{2-5}
         & \textbf{TRUE} & \textbf{FALSE} & \textbf{TRUE} & \textbf{FALSE} \\ \hline
        \textbf{Enrolled Reference} & TM & \cellcolor[HTML]{FFE2C4} FNM  & TM & \cellcolor[HTML]{FFE2C4} FNM \\ \hline
        \textbf{Non-Enrolled Reference} & \cellcolor[HTML]{FFE2C4} EFM $^a$ & N/A $^b$ & \cellcolor[HTML]{FFE2C4} EFM/UFM $^c$ & TNM \\ 
        \hline
        \multicolumn{5}{l}{%
            \begin{minipage}{\linewidth}%
                \setstretch{0.6}
                \footnotesize
                $^a$ \textit{Enrolled False Match}: An enrolled non-mated biometric reference is similar enough to be considered a match. \\
                $^b$ True Non-Match cannot happen, because there are no unenrolled references. \\
                $^c$ \textit{Unenrolled False Match}: An unenrolled biometric reference is similar enough to be considered a match.
            \end{minipage}%
        } 
        \end{tabular}
    }
\end{table}

\subsection{Traditional searching: \emph{1:N}}
\label{sec:1Nsearch}


\citet{Mukherjee2008} define the problem of iris identification in terms of comparing a probe iris sample $q$, with enrolled iris samples $D=\{d_1,d_2,d_3,...d_n\}$, in order to determine the identity $y$ of the query sample.
Each enrollment sample $d_j$, $j=1,2,...,n$ is associated with an identity $y_j$.
Consequently, the computational complexity of the process is directly linked to the number of enrolled biometric references $|D|=n$ in the enrollment.

Matching iris samples based on Daugman's approach is an operation that involves the accumulation of bitwise XOR operations between the biometric templates, and can be done quite efficiently.
However, the computational complexity of the task grows linearly with the increase in enrollment size, and the complexity for tasks like de-duplication grows quadratically regarding the size of the database, as noted by \cite{Proenca2013}.
%
%

Unlike other numeric or lexicographic data, biometric samples do not have any natural ordering \cite{Rathgeb2010}. 
This hinders any attempt to index biometric databases.
Since there is no order for the enrollment records, the obvious approach used in automatic iris identification is to compare the probe to every enrollment record.

Other efforts have been made in the sense of improving the search performance in iris databases. 
In \cite{rakvic2009}, the parallelization of the algorithms involved in the iris recognition process is proposed, including the template matching.  
However, their parallelized version still has its overall performance directly associated to the size of the database.
In another attempt to address the issue, \citet{hao2008} propose an approach based on Nearest Neighbor search, to reduce the search range and thus improve the performance. 

If the number of biometric references to be searched is large, it may be slow to sweep it entirely every time a user is presented to the system.  
The traditional approach for the implementation of one-to-many identification is the exhaustive 1:N, and is probably the only form of one-to-many search to receive attention in the research literature.

\subsection{The alternatives and \emph{1:First search}}
\label{sec:1Firstsearch}

The need for optimization of computational requirements is motivated by the large scale of many iris recognition deployments, together with a tendency for deployments to grow larger over time.
Naturally, one can expect the computational demands to increase.
Nevertheless, only a small number of proposed indexing methods are found in the literature \cite{Proenca2013}.

A relatively common approach to speed up the search process is to perform a search known as \emph{1:First}, in which the system sweeps the biometric enrollment until it finds the first template for which the comparison score meets a defined threshold, and declares a match \citep{Ortiz2015}. 
This approach yields a lower number of comparisons on average, compared with the 1:N method. 
On the other hand, it may lead to a higher error rate, since when a match pair is found the search is stopped, ignoring other potentially better matches.

If the biometric references are randomly distributed in the database, it is likely that acceptable matches will be found before the end of the list, thus reducing the search time. However, this approach raises some questions which have been only partially answered so far: 
\begin{enumerate}
    \item Does 1:First result in the same identification accuracy as 1:N?
    \item If not, how frequently does an acceptable match correspond to a correct match (the biometric probe and the biometric reference correspond to the same biometric subject)?
    \item How good does a match have to be to be considered acceptable? 
    \item Empirically, does 1:First perform faster than 1:N? If so, how much faster?
\end{enumerate}


\citet{Bowyer2013survey} surveyed several different attempts to improve the matching speed of iris codes and reduce the time required for database scans.
These works report different degrees of success, but there seems to be no clear trend in performance for iris matching and searching.
Furthermore, to the best of our knowledge, \citet{kuehlkamp2016_1first} presented the only work to have experimentally evaluated the 1:First search technique.

\section{Methods}
\label{sec:methods}


Previous results from \citep{kuehlkamp2016_1first} suggested that the FMR grows as the enrollment size is increased.
However, limitations of this work prevent a better understanding of the global scenario of 1:First performance: enrollment sizes in the range of 100--1400 subjects are hardly representative of real world applications; only closed-set scenarios were tested; a single academic iris matcher was employed; and finally, only a single ordering of the biometric references was used in the experiments.

A larger and more complex set of experiments was conceived to address these issues: 1) number of biometric references belonging to different subjects in an enrollment ranging from 500 to 11,000; 2) closed- and open-set scenarios; 3) use of a commercial matcher in addition to IrisBee for comparison; 4) comparing a biometric probe set against multiple random permutations of the same enrollment of biometric references.

\subsection{Rotation Tolerance}
\label{sec:rottol}

Rotational mis-alignment can cause two images of the same iris to be misinterpreted as a non-match.
For this reason, iris matchers implement tolerance to iris rotation, in the form of ``best-of-M'' comparisons: A pair of iris codes is compared several times, over a range of relative rotations, and the best match is chosen to be the comparison score for that pair \citep{Daugman2006}. 
As an example, the CBSA NEXUS border-crossing program considers a range of 14 rotation values in the initial scan of the biometric enrollment, and the range is widened to an additional 28 rotation values if no match was found on the initial scan \citep{Chumakov1}.
This results in a faster initial scan that picks up the large majority of rotational mis-alignments, and a further scan, only when needed, to pick up more extreme mis-alignments.

\subsection{Closed-set \emph{versus} Open-set}

In \emph{closed-set identification} the system tries to determine the identity of an unidentified individual whom is known to be \emph{enrolled} in the database.
Conversely, in \emph{open-set identification}, it is not known if the individual presenting a probe sample is enrolled in the database. 
In this case, the system has to determine if the user is in the database, and if so, to find the corresponding enrolled identity.

The fundamental performance metrics for matching are False Match Rate (FMR) and False Non-Match Rate (FNMR), as defined by \citet{ISO19795}. 
Since our evaluation is made at the algorithm level, we do not take into account acquisition or enrollment failures. 
FMR is the proportion of probes that are incorrectly declared to match the enrollment of some different person, while FNMR is defined as the proportion of probes that are incorrectly declared to not match the enrollment of the same person.
In their more specific sense, FMR and FNMR refer to outcomes of 1:1 matching, although the terms are often also informally used in the 1:N context.

The standard \citep{ISO19795} also defines metrics specific for the identification (1:N matching) task, namely: True-Positive Identification Rate (TPIR or simply IR), False Negative Identification Rate (FNIR) and False Positive Identification Rate (FPIR). 
In general, TPIR, FNIR and FPIR are terms in the 1:N matching scenario that correspond to the terms TMR, FNMR, and FPR in the 1:1 matching scenario.

FPIR is a metric that applies specifically to open-set contexts, since it refers to non-enrolled user presentations, which do not occur in closed-sets.
However, there is no definition in the standard for operations in which biometric probes are incorrectly matched to some wrong biometric reference (of a different subject).
We use the term \textit{Enrolled False Positive Identification Rate} (E-FPIR) as an adaptation to extend the open-set metric to cases which, in verification scenario would be comprehended under FMR.
 
In our experiments, an identification transaction is the presentation of one probe to the system, which results in one of two possible outcomes: a) an identity label assigned to that probe, or b) no identity assigned to the probe, implying that the probe does not match any of the biometric references.
The correctness of the result falls into one of five possible categories: 
\begin{enumerate}
    \item{True Identification (TI)} -- occurs when the system returns a biometric reference identifier that corresponds to the biometric subject whose probe was presented;
    \item{Enrolled False Positive Identification (E-FPI)} -- the system returns a biometric reference identifier that does not correspond to the biometric subject who presented the probe, and corresponds to another biometric subject whose reference is stored in the database;
    \item{True Non-Identification (TNI)} -- the system returns no biometric reference identifier and the corresponding biometric reference was not enrolled (this situation can only happen in an open-set scenario);
    \item{False Non-Identification (FNI)} -- the system returns no biometric reference identifier, but the corresponding biometric reference was in the database;
    \item{False Positive Identification (FPI)} -- the system returned a biometric reference identifier, but the presented probe does not correspond to any biometric references (this situation only occurs in an open-set scenario).
\end{enumerate}

Using these result categories, we calculate the accuracy metrics in terms of E-FPIR, FPIR and FNIR for all experiments, in order to evaluate accuracy in each category. 
In closed-set experiments, FPIR is omitted because it is, by definition, zero.

\subsection{Dataset}
\label{sec:dataset}


Images were selected from a dataset captured using an LG-4000 sensor, during acquisition sessions performed over the years of 2008 through 2013 at the University of Notre Dame. 
In addition to the 57,232 images used in the experiments presented in \citep{kuehlkamp2016_1first}, the University's repository contains 51,234 images captured with the same sensor, after a firmware update.
Thus, a total of 108,466 images of left and right eyes were selected, representing a total of 1,991 people, and 3,982 individual eyes.

Before combining the images from before and after the firmware update into one pool for our experiments, it was necessary to analyze the two sets of images to ensure the update did not introduce changes that could interfere with the results.
Four different aspects of the two sets of images were compared: a) intensity distributions; b) comparison score distributions; c) image count distribution by subject; and d) image focus.
While intensity and focus did not reveal significant changes after the upgrade, the analysis of comparison score and subject distributions revealed some variation. 
An examination of the data revealed the distortions were caused not because of the firmware update, but because of a few subjects who had a much higher image count than the average due to participating in experiments for specific research purposes. 
These subjects were removed from this dataset.

\subsubsection{Simulating more individuals for increased enrollment size}

In order to perform experiments on the largest possible galleries and probe sets, we use a data augmentation technique to increase the number of unique eyes in our dataset.
It is known that there is no correlation between the left and right iris of the same person, and in the same sense, orientation of two different images of the same iris must be correctly aligned in order to generate an identity match \citep{Czajka2017_LRUD}.
Based on this, we performed two spatial transformations on the original images of the set: 180$^{\circ}$ rotation and horizontal flipping.

\begin{figure}[!b]
    \centering
    \includegraphics[width=0.7\textwidth]{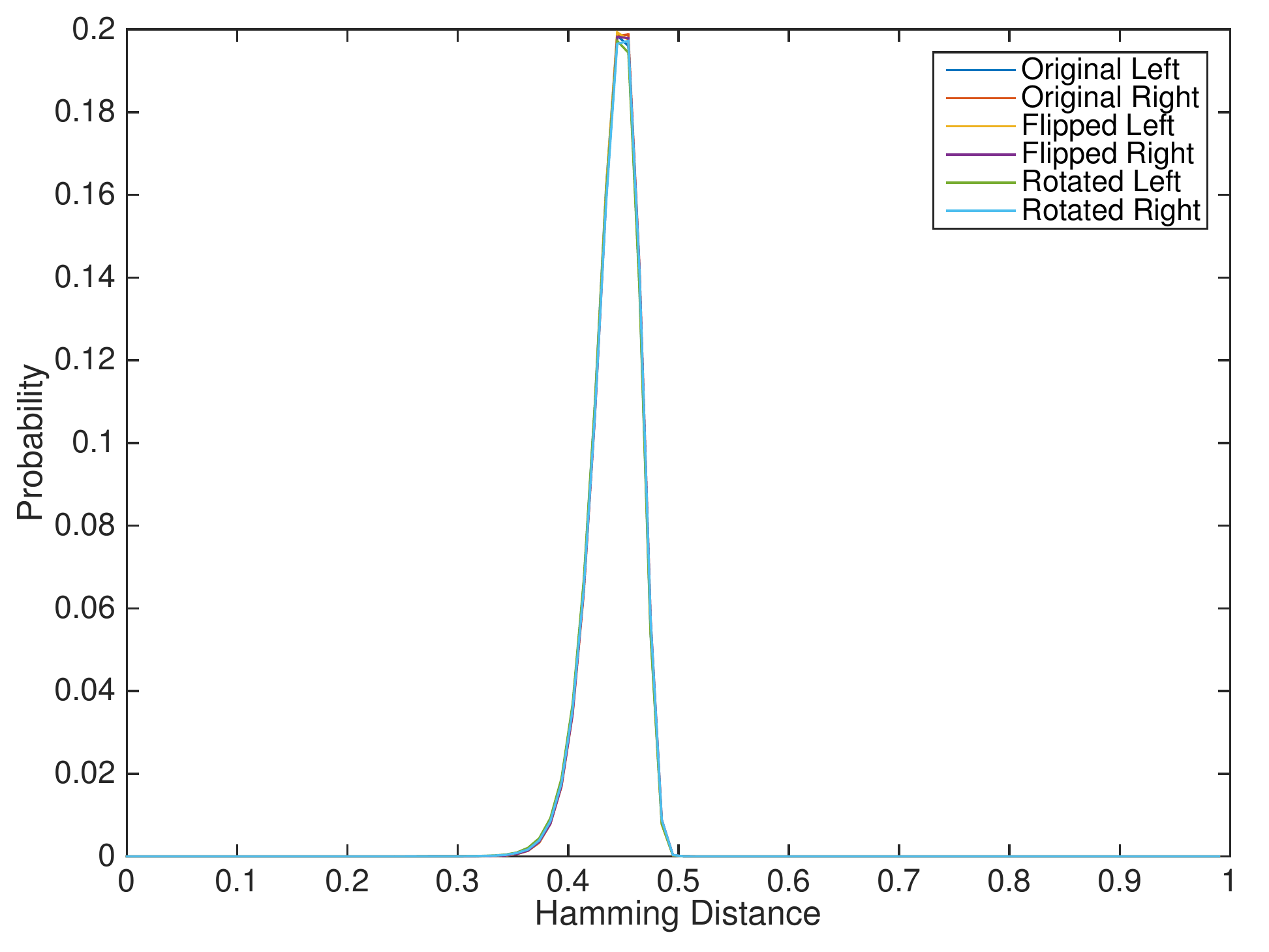}
    \caption{Comparison of HD distributions between original and artificial images.}
    \label{fig:distr_artificial}
\end{figure}

To make sure the spatial transformation does not result in a set that diverges from the properties of the original, we selected the oldest image of each eye of each subject, on which we applied rotation and flipping, creating four other sets of 1,991 images each.
Next, we performed all-versus-all matching using IrisBee and compared the Hamming Distance distributions. 
Each set of images resulted in approximately 2 million one-way comparisons.
As shown in Figure \ref{fig:distr_artificial}, there is very little difference between the six sets.

\begin{figure}[!htbp]
    \centering
    \includegraphics[width=0.7\textwidth]{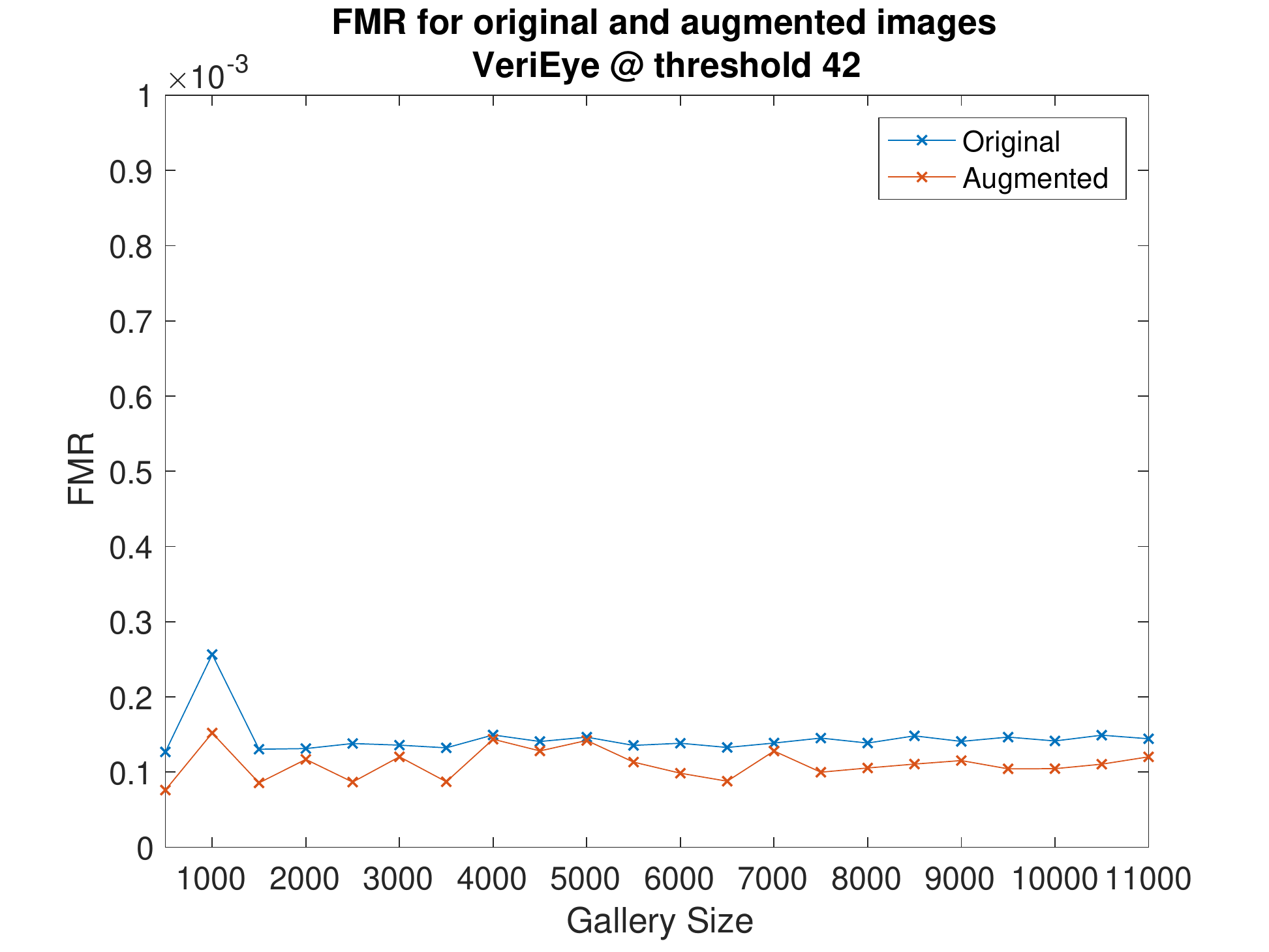}
    \caption{Comparison of FMR for matching performed on original and artificial images. In all cases, FMR is below 0.0003.}
    \label{fig:fmr_artificial}
\end{figure}

As an additional verification, we compared the FMR obtained from matching original images and from matching artificial images. 
Abnormalities in the images or the matching process would likely cause an increase in FM, however that did not happen.
Figure \ref{fig:fmr_artificial} illustrates this: FMR is slightly lower for artificial images than it is for original images. 
This is justified by the fact that there are approximately three times more artificial images than original ones.

The final data set is composed by the union of these six sets, amounting to 11,946 unique eyes, which for the purposes of this work, were then considered as unique subjects.
Finally, the rotation and flipping transformations were applied to all available images, resulting in a total of 325,398 images that were used in the experiments.

\subsection{Enrollment and Probe Set Formation}
\label{sec:gall_probesets}


From this augmented data set, we created 22 subject-disjoint galleries varying in size from 500 to 11,000 subjects, in increments of 500. 
To select these galleries, we picked the single oldest image for each person, for each eye, and for each transformation (original, rotated and flipped).
We call this set the \textit{Enrollment Pool}.
Images were randomly drawn from the Enrollment Pool to form each of the experimental galleries.

After forming the galleries, the remaining 313,452 images were used as a \textit{Probe Pool} to create closed probe sets for each of the galleries.
Then, for each subject in the enrollment, the corresponding images in the Probe Pool were added to the probe set.
The total size of each probe set varies according to the number of images available for each subject. 
The smallest probe set has 10,836 probes, and the largest was limited to 20,000.

To create open probe sets, a similar procedure was adopted: for each enrollment subject, a closed probe set with size $N \times 1.5$ was randomly drawn from the Probe Pool, where N is the size of the enrollment. 
Next, a set of $\frac{N \times 1.5}{2}$ images from subjects not in the enrollment was added to the selection to form the open portion of the probe set.

Finally, for each image the corresponding biometric references and biometric feature sets (constituting biometric probes) were calculated using two different iris recognition methods: IrisBee and Neurotechnology VeriEye. 

As an example, an enrollment of 500 subjects yields a closed probe set of biometric probes corresponding to $500 \times 1.5 = 750$ enrolled subjects, and biometric probes corresponding to $750 / 2 = 375$ non-enrolled subjects are added as the open (or non-enrolled) portion of the probe set. The final probe set size in this case is $750 + 375 = 1,125$ probes, where $\sim 33\%$ of them correspond to non-enrolled subjects.
Table \ref{tab:galProbeCompo} describes the composition of the galleries and their respective probe sets.

\begin{table}[t]
\centering
\caption{Enrollment and Probe Set Composition}
\label{tab:galProbeCompo}
    \resizebox{\linewidth}{!}{%
        \begin{tabular}{r|r|r|r|r|r|r|r|r|r|r|r}
        \hline
        \multicolumn{4}{c|}{\textbf{Enrollment}} & \multicolumn{4}{c|}{\textbf{Closed Probe Set}} & \multicolumn{4}{c}{\textbf{Open Probe Set}} \\ \hline
        \textbf{Size} & \textbf{Original} & \textbf{Rotated} & \textbf{Flipped} & \textbf{Size} & \textbf{Original} & \textbf{Rotated} & \textbf{Flipped} & \textbf{Size} & \textbf{Original} & \textbf{Rotated} & \textbf{Flipped} \\ \hline
        \textbf{500} & 167 & 163 & 170 & \textbf{10836} & 3303 & 4222 & 3311 & \textbf{1125} & 388 & 366 & 371 \\
        \textbf{1000} & 327 & 342 & 331 & \textbf{20000} & 6679 & 6004 & 7317 & \textbf{2250} & 730 & 752 & 768 \\
        \textbf{1500} & 474 & 507 & 519 & \textbf{20000} & 6502 & 6180 & 7318 & \textbf{3375} & 1111 & 1103 & 1161 \\
        \textbf{2000} & 661 & 667 & 672 & \textbf{20000} & 6751 & 6678 & 6571 & \textbf{4500} & 1524 & 1489 & 1487 \\
        \textbf{2500} & 806 & 848 & 846 & \textbf{20000} & 5950 & 6928 & 7122 & \textbf{5625} & 1832 & 1912 & 1881 \\
        \textbf{3000} & 1024 & 995 & 981 & \textbf{20000} & 6570 & 6822 & 6608 & \textbf{6750} & 2306 & 2219 & 2225 \\
        \textbf{3500} & 1184 & 1163 & 1153 & \textbf{20000} & 6850 & 6090 & 7060 & \textbf{7875} & 2658 & 2644 & 2573 \\
        \textbf{4000} & 1298 & 1340 & 1362 & \textbf{20000} & 6714 & 6363 & 6923 & \textbf{9000} & 3015 & 2999 & 2986 \\
        \textbf{4500} & 1493 & 1513 & 1494 & \textbf{20000} & 6530 & 6994 & 6476 & \textbf{10125} & 3415 & 3397 & 3313 \\
        \textbf{5000} & 1675 & 1654 & 1671 & \textbf{20000} & 6554 & 6846 & 6600 & \textbf{11250} & 3878 & 3705 & 3667 \\
        \textbf{5500} & 1836 & 1858 & 1806 & \textbf{20000} & 6984 & 6672 & 6344 & \textbf{12375} & 4129 & 4131 & 4115 \\
        \textbf{6000} & 2012 & 1986 & 2002 & \textbf{20000} & 6238 & 6899 & 6863 & \textbf{13500} & 4532 & 4423 & 4545 \\
        \textbf{6500} & 2162 & 2199 & 2139 & \textbf{20000} & 6823 & 6422 & 6755 & \textbf{14625} & 4853 & 5005 & 4767 \\
        \textbf{7000} & 2355 & 2318 & 2327 & \textbf{20000} & 6802 & 6466 & 6732 & \textbf{15750} & 5183 & 5264 & 5303 \\
        \textbf{7500} & 2482 & 2488 & 2530 & \textbf{20000} & 6715 & 6667 & 6618 & \textbf{16875} & 5557 & 5644 & 5674 \\
        \textbf{8000} & 2730 & 2607 & 2663 & \textbf{20000} & 6808 & 6389 & 6803 & \textbf{18000} & 6014 & 6058 & 5928 \\
        \textbf{8500} & 2817 & 2845 & 2838 & \textbf{20000} & 6740 & 6805 & 6455 & \textbf{19125} & 6371 & 6301 & 6453 \\
        \textbf{9000} & 2988 & 3023 & 2989 & \textbf{20000} & 6651 & 6689 & 6660 & \textbf{20250} & 6776 & 6829 & 6645 \\
        \textbf{9500} & 3147 & 3182 & 3171 & \textbf{20000} & 6641 & 6768 & 6591 & \textbf{21375} & 7108 & 7058 & 7209 \\
        \textbf{10000} & 3365 & 3328 & 3307 & \textbf{20000} & 6726 & 6797 & 6477 & \textbf{22500} & 7514 & 7328 & 7658 \\
        \textbf{10500} & 3502 & 3535 & 3463 & \textbf{20000} & 6657 & 6739 & 6604 & \textbf{23625} & 7685 & 7722 & 8218 \\
        \textbf{11000} & 3674 & 3676 & 3650 & \textbf{20000} & 6723 & 6638 & 6639 & \textbf{24750} & 8140 & 8277 & 8333 \\ \hline
        \end{tabular}
    }
\end{table}

\subsection{Threshold Selection}
\label{sec:thressel}


The ``strictness" of an iris recognition application is regulated by a threshold that stipulates the minimum similarity required between two samples so that they can be considered a match.
This threshold is usually set experimentally, and it is defined as a value in the scale of the matcher output.

Daugman-style iris matchers like IrisBee use Hamming Distance as the scale of dissimilarity between samples. 
This scale goes from 0.0 (no dissimilarity) to 1.0 (complete dissimilarity). 
VeriEye, on the other hand, uses a similarity scale to compare samples.
It ranges from 0 (minimal similarity) to a maximum observed similarity score of 9,443. 
Since there is no direct relation between the two scales, it is necessary to establish an equivalence between values, so that we can make a fair comparison of the two matcher results.

\begin{figure}[!tb]
    \centering
    \includegraphics[width=0.33\textwidth]{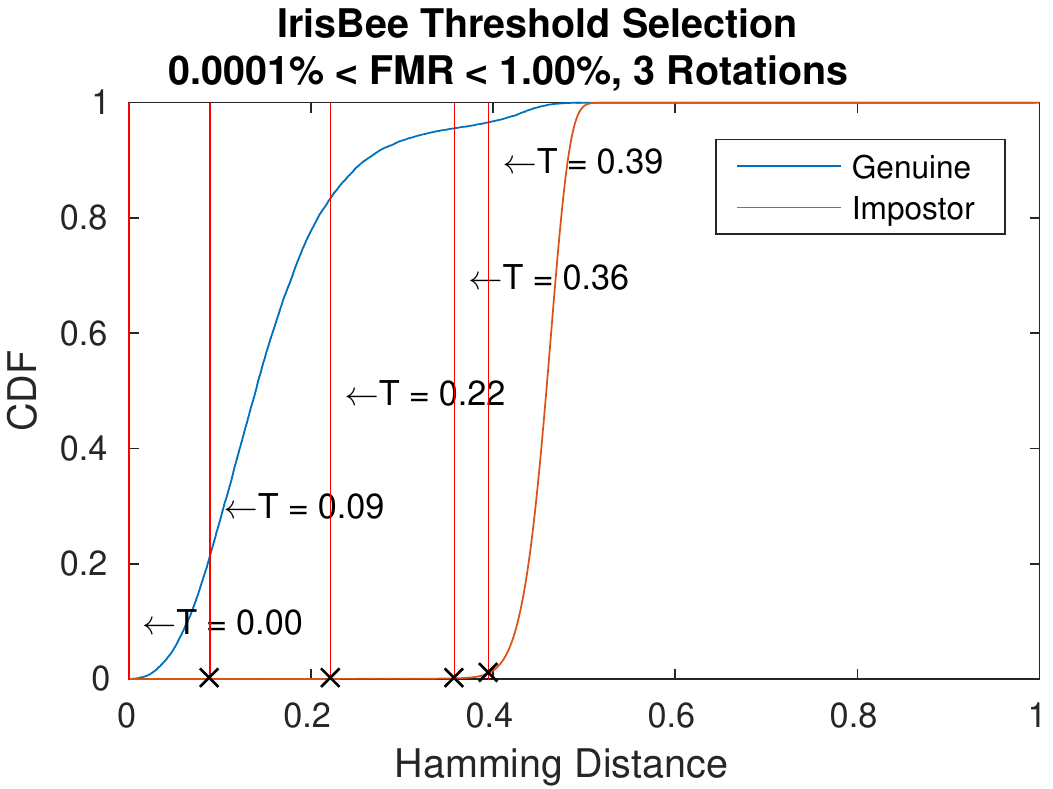}\hfill%
    \includegraphics[width=0.33\textwidth]{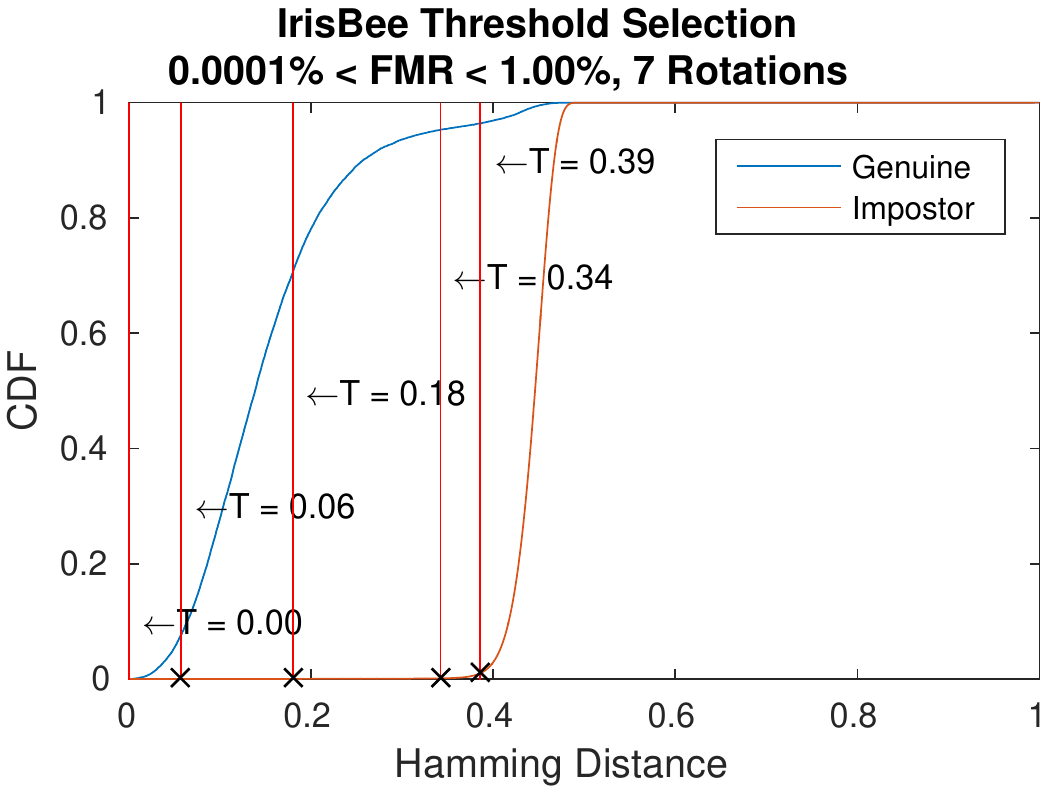}\hfill%
    \includegraphics[width=0.33\textwidth]{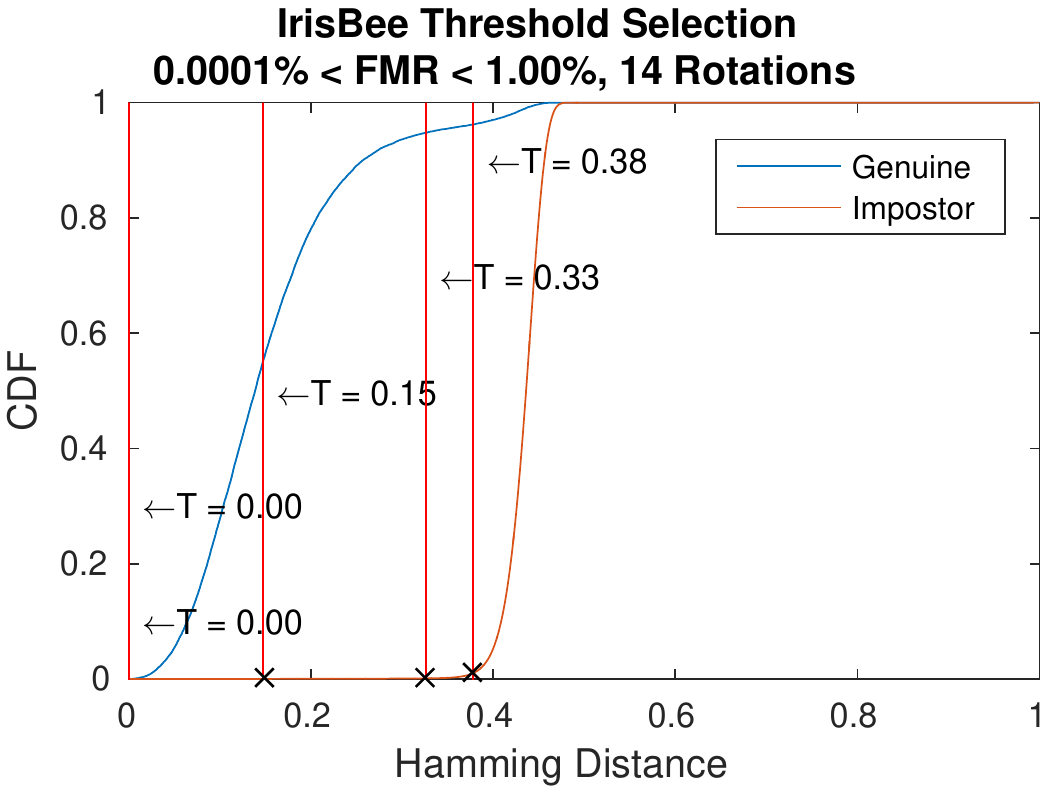}\hfill%
    \\
    \includegraphics[width=0.33\textwidth]{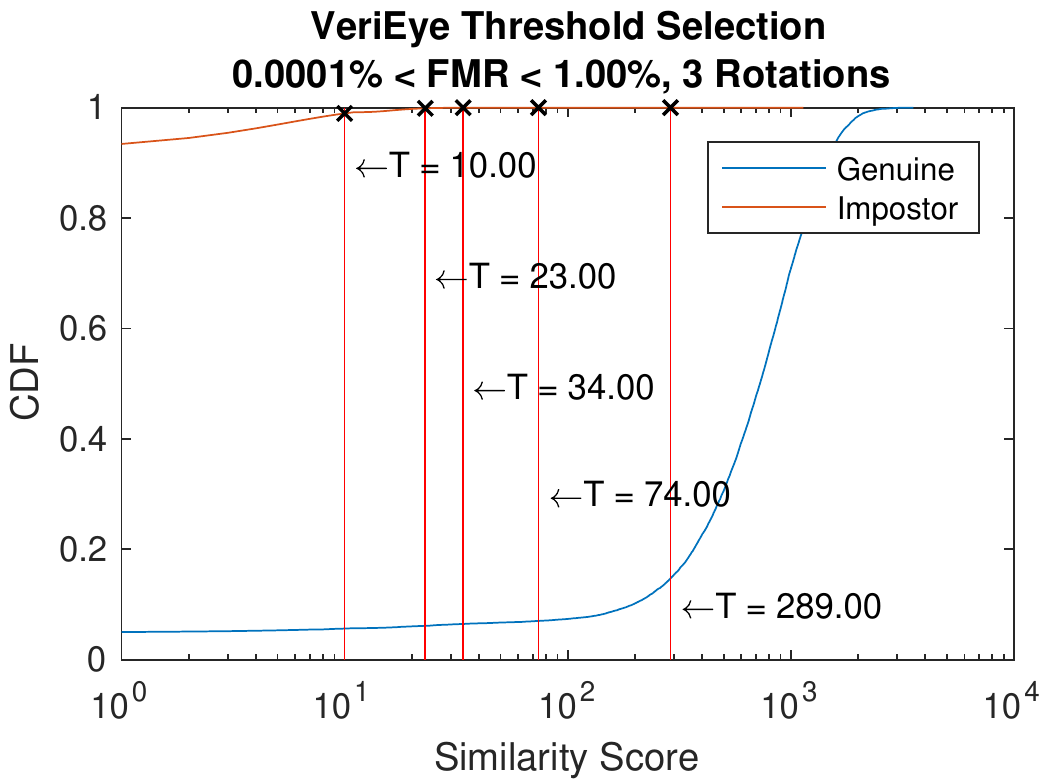}\hfill%
    \includegraphics[width=0.33\textwidth]{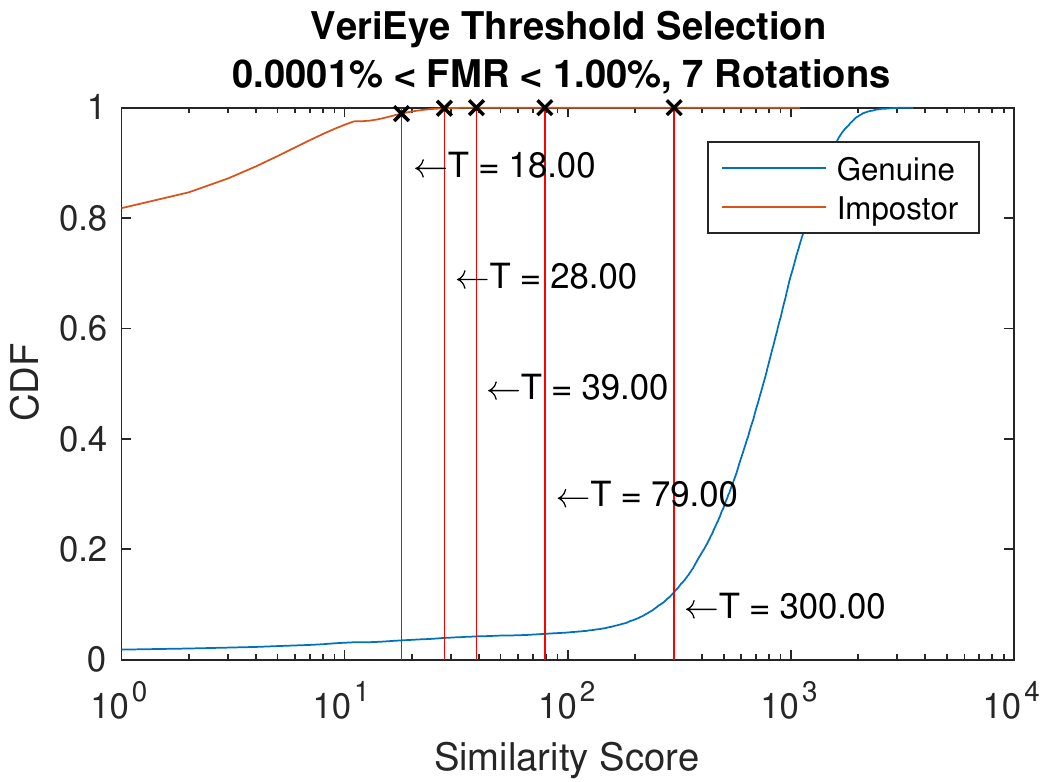}\hfill%
    \includegraphics[width=0.33\textwidth]{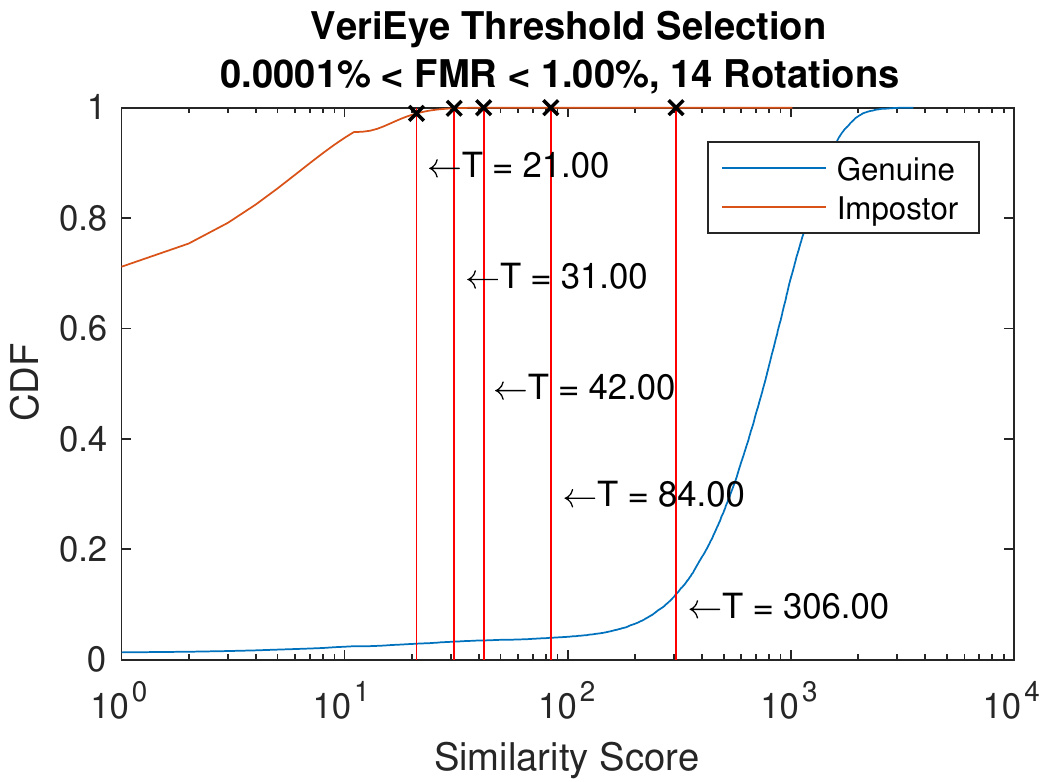}\hfill%
    \caption{CDF-based threshold selection. Observe that the Cumulative Distribution curves are very different in format for IrisBee (top row) and VeriEye (bottom row): This happens because the matchers use dissimilarity and similarity scales, respectively.}
    \label{fig:thres_sel}
\end{figure}

To do so, we ran matching using the largest available enrollment (11,000 references) and its corresponding probe set (20,000 probes), resulting in more than 200 million comparisons. 
Using the comparison score output for each comparison, we plotted the Cumulative Distribution Function (CDF) for the genuine and impostor distributions.
Figure \ref{fig:thres_sel} shows examples of the threshold selection for IrisBee and VeriEye.
Based on the impostor CDFs, five threshold values were selected, corresponding to $0.0001\%$, $0.001\%$, $0.01\%$, $0.1\%$ and $1\%$ of the impostor comparisons.
This ensures we have thresholds corresponding to five distinct \textit{Accuracy Targets}. An example of this equivalence is shown in Table \ref{tab:cdf_thres}.

\begin{table}[!h]
	\caption{Empirically selected threshold for closed-set with 14 rotations}
	\label{tab:cdf_thres}
	\begin{center}
    \resizebox{\columnwidth}{!}{
		\begin{tabular}{ c | c | c }
			\hline
			\textbf{Accuracy Target (Max. False Matches)} & \textbf{IrisBee Threshold} & \textbf{VeriEye Threshold} \\ \hline
			1\% &   0.3780 &    21 \\ \hline
			0.1\% &   0.3263 &    31 \\ \hline
			0.01\% &   0.1475 &    42 \\ \hline
			0.001\% &   0{$^a$} &    84 \\ \hline
			0.0001\% &   0{$^a$} &    306 \\ \hline
            \multicolumn{3}{l}{%
                \begin{minipage}{\linewidth}%
                    \setstretch{0.6}
                    \footnotesize
                    \item{$^a$} The matcher could not achieve the accuracy target.
                \end{minipage}%
            }\ 
		\end{tabular}
	}
	\end{center}
\end{table}

\section{Results and Discussion}
\label{sec:results}

\subsection{Closed-set Scenario}

\begin{figure}[!t]
    \centering
    \includegraphics[width=\textwidth]{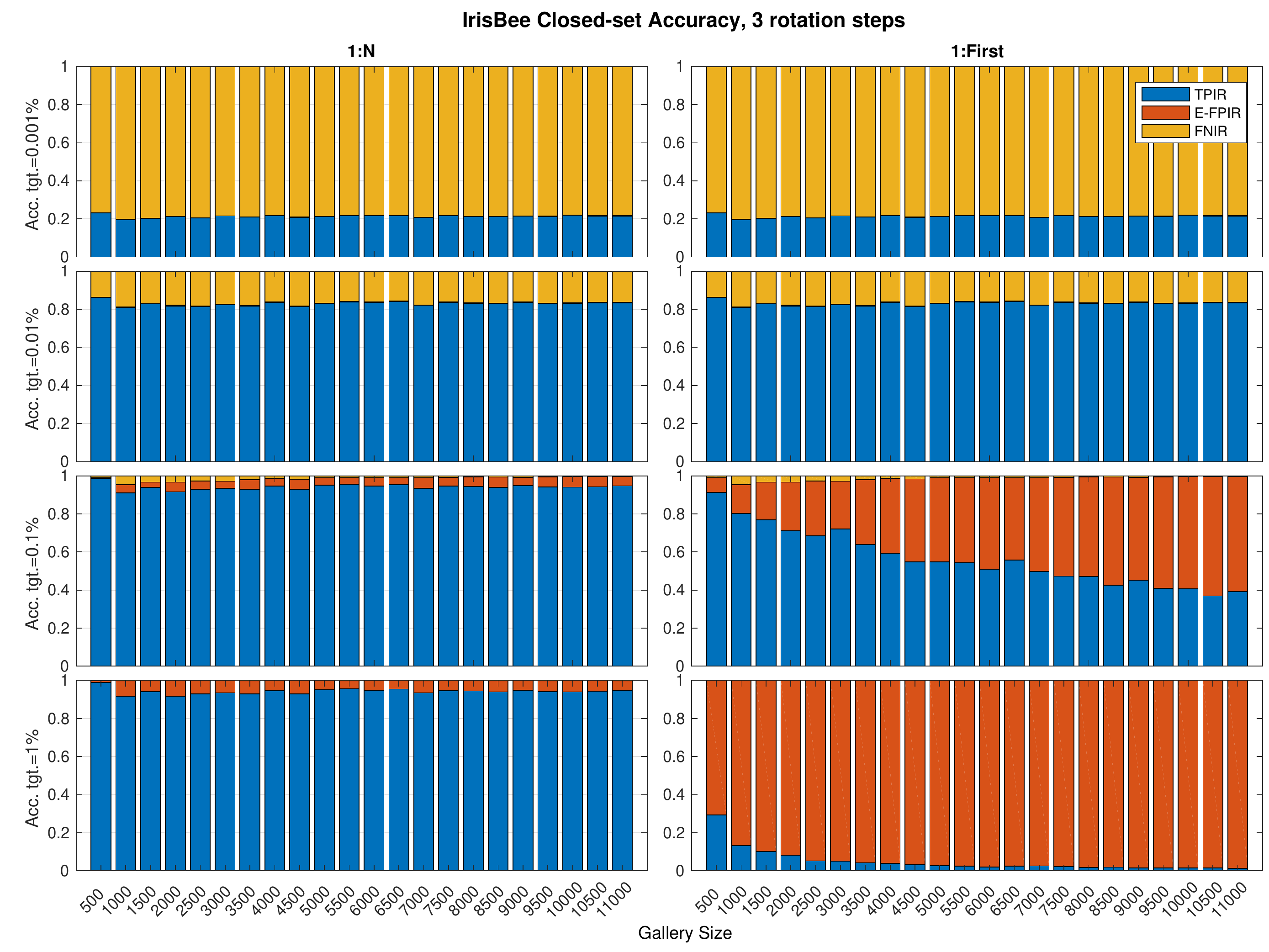}
    \caption{Identification performance of IrisBee in closed-set scenario, at different accuracy targets.}
    \label{fig:acc_ib_closed_thr}
\end{figure}

\begin{figure}[!t]
    \centering
    \includegraphics[width=\textwidth]{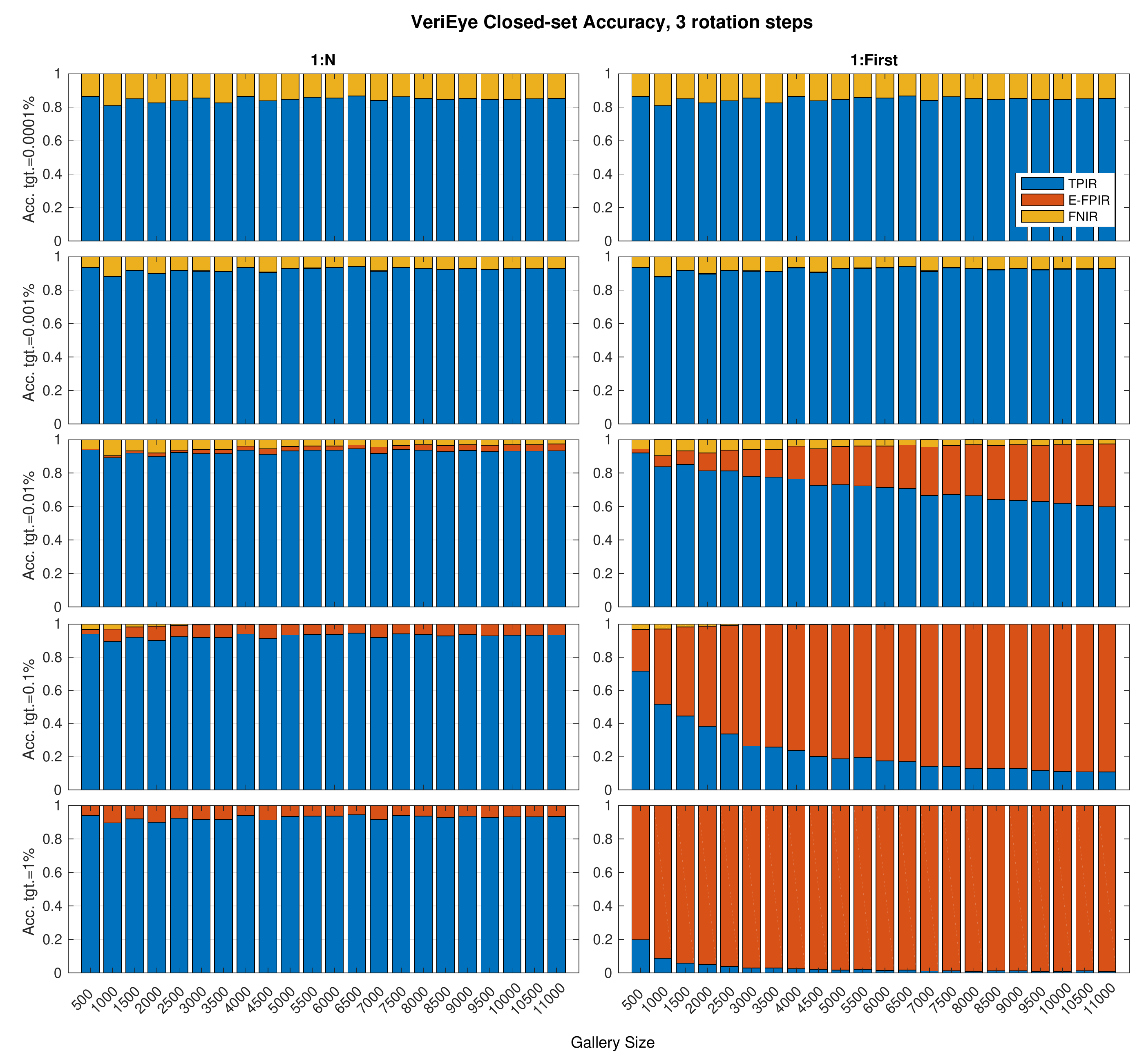}
    \caption{Identification performance of VeriEye in closed-set, at different accuracy targets.}
    \label{fig:acc_ve_closed_thr}
\end{figure}

Figures \ref{fig:acc_ib_closed_thr} and \ref{fig:acc_ve_closed_thr} show a performance comparison between 1:N and 1:First, in terms of their performance scores, in closed-set scenarios under different accuracy settings. 
Identification was performed using both matchers, over a range of accuracy targets going from more strict (top) to less strict (bottom).
In the case of the less strict accuracy targets, we can perceive a pronounced trend in 1:First (right column) performance: E-FPIR grows continuously as the enrollment size increases, but the same does not occur in 1:N (left column).
In the worst cases (larger galleries with less strict accuracy target), 1:First E-FPIR reaches nearly 100\%, while the same metric obtained by 1:N remains below 10\%.

Comparison between Figures \ref{fig:acc_ib_closed_thr} and \ref{fig:acc_ve_closed_thr} confirms the expected higher accuracy of commercial matcher VeriEye as opposed to the research software IrisBee. 
The most strict accuracy target achieved by IrisBee is 0.001\% with an FNIR around 80\%, while VeriEye reached a target accuracy of 0.0001\% with FNIR not going above 20\%.
Both matchers, however, show the same types of trends when we compare 1:N and 1:First searches. 
1:First search shows a strong tendency to higher E-FPIRs when the target accuracy is less strict than 0.01\%.
On the other hand, if the target accuracy is more strict than a certain limit, 1:N and 1:First are very similar. 
As it could be expected given the differences in overall accuracy between the matchers, this threshold is different for each of them: with IrisBee, 1:N and 1:First have similar performances for target accuracies more strict than 0.01\%.
With VeriEye, the same occurs when target accuracy is more strict than 0.001\%.
This is verified by the similarly sized TPIR and FNIR bars in the upper plots in these figures.

FNIR however does not suffer the same kind of influence by the search method: both IrisBee and VeriEye display the same values for FNIR, regardless if they were achieved through 1:N or 1:First search.
This is explained by the definition of FNIR.
In order for a \textit{Non-Match} to be declared, it is necessary that the system have gone through the entire enrollment. 
In this case, there is no difference between 1:N and 1:First search.

\subsection{Rotation Tolerance}

\begin{figure}[!htb]
    \centering
    \includegraphics[width=\textwidth]{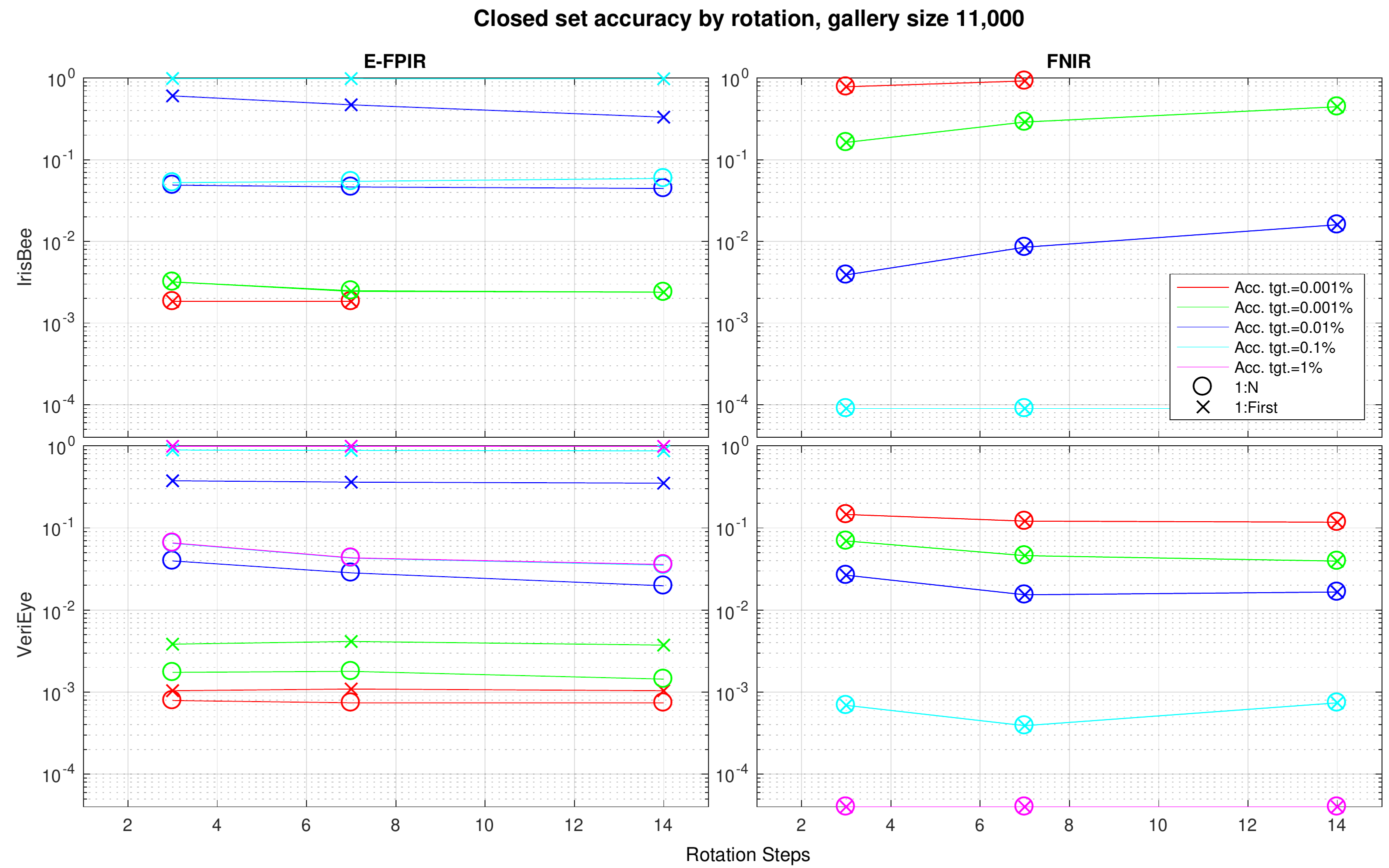}
    \caption{Identification performance across different rotation tolerances.}
    \label{fig:acc_closed_rot}
\end{figure}

Tolerance to iris rotation can interfere with the accuracy of the matcher, since the higher the number of rotations allowed, the higher the chance that the IrisCode will randomly match another unrelated iris.
This interference can be compensated by the selection of a more restrictive threshold, as the rotational tolerance is increased.
In the threshold selection procedure described in Section \ref{sec:thressel}, we have considered three rotational tolerance levels and chosen the thresholds individually to compensate for it.
To illustrate this, Figure \ref{fig:acc_closed_rot} demonstrates a very small variance in E-FPIR and FNIR across different rotational tolerances, in both matchers and search methods.

\subsection{Open-set Scenario}

\begin{figure}[!t]
    \centering
    \begin{subfigure}[b]{\textwidth}
        \centering
        \includegraphics[width=\textwidth]{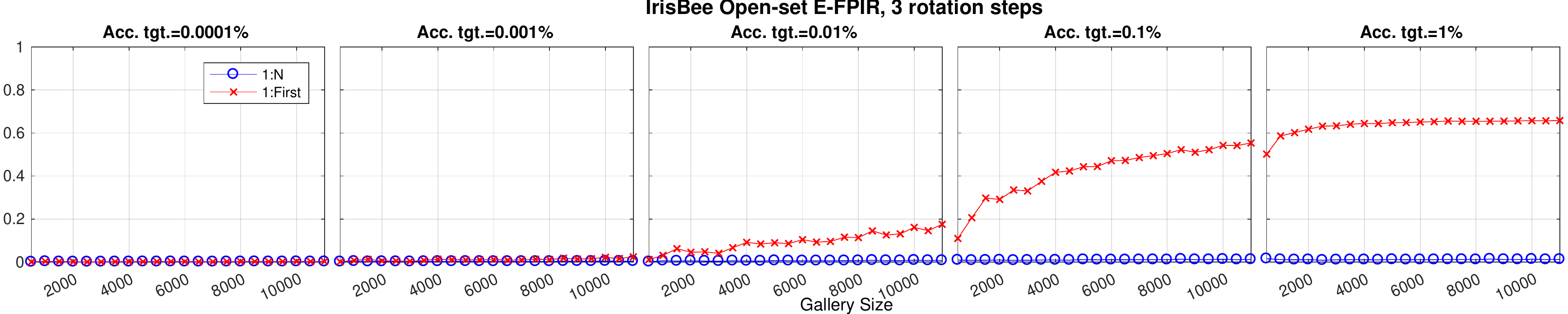}
        \caption{IrisBee.}
        \label{fig:acc_ib_open_thr}
    \end{subfigure}
    
    \begin{subfigure}[d]{\textwidth}
        \centering
        \includegraphics[width=\textwidth]{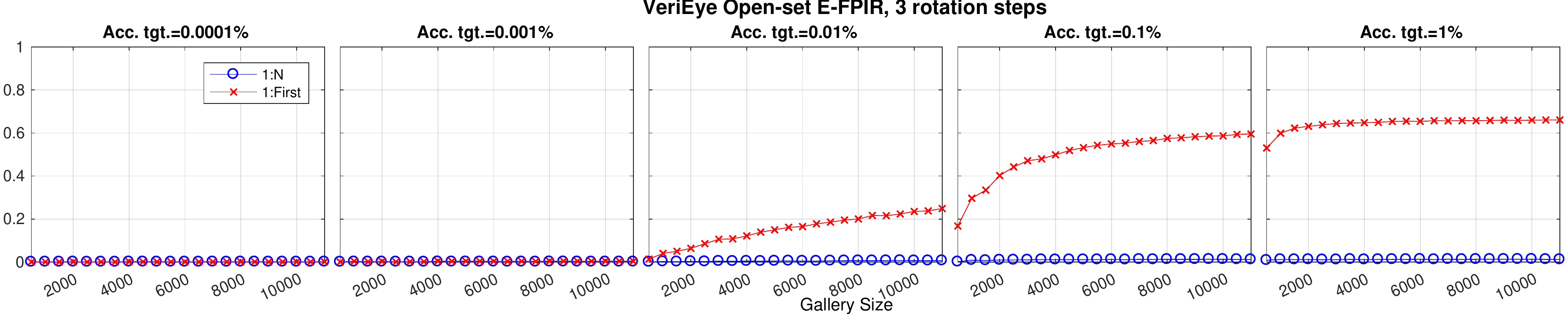}
        \caption{VeriEye}
        \label{fig:acc_ve_open_thr}
    \end{subfigure}
    
    \caption{E-FPIR identification performance in open-set scenario, at different accuracy targets.}
    
\end{figure}

Open-set scenario results can be seen in Figures \ref{fig:acc_ib_open_thr} and \ref{fig:acc_ve_open_thr}.
In these cases, $33.33\%$ of the probes correspond to subjects that are not enrolled in the enrollment.
In these scenarios, we only present results in terms of E-FPIR, since there was no measurable difference between 1:N and 1:First in FPIR, FNIR and TPIR.
Also, we do not use the same type of bar plot, because E-FPIR and FPIR are not complementary i.e., their summation exceeds 100\%.
The same types of trends found in the closed-set scenarios are present here: if the accuracy target of the system is not highly restrictive, 1:First search yields worse results than 1:N, and the difference grows proportionally to the size of the galleries.
At first, comparing the worst E-FPIR open-set cases with the corresponding closed-set experiment, E-FPIR seems to have decreased (in some cases, from $\sim100\%$ to $\sim60\%$). 
This fact is explained by the presence of unenrolled subjects in the probe set, which corresponds to one third of the probes.

An interesting note about this is that despite the fact that VeriEye is in general more accurate than IrisBee, E-FPIR behavior manifested similarly in both matchers. 
At accuracy target 0.01\%, E-FPIR is generally higher for VeriEye than for IrisBee, when using 1:First search (red series in Figs. \ref{fig:acc_ib_open_thr} and \ref{fig:acc_ve_open_thr}). 

Although not shown in the figures, another interesting fact revealed in the open-set scenario is that the negative impact of 1:First search in E-FPIR is not verified in FPIR.
Again, it can be explained if we look closely at what happens in 1:First search: if a search for an enrolled probe is interrupted because a better-than-threshold match was found, there is a non-zero probability that a better (and true) match could be found if the search was continued -- that is, if we were performing 1:N instead, we could get a better result by finding the genuine match (and consequently declaring a true match).
But if the search for a probe that is \textit{not enrolled} is interrupted in 1:First, the best that we could do if we continue the search is to find a better false match, so the result would still be a false match.

Like in closed-set experiments, there is practically no difference between 1:N and 1:First scores at highly strict accuracy targets (0.0001\%, on the left of the figure).
As the system strictness is relaxed, 1:First E-FPIR starts to grow much higher than 1:N E-FPIR, increasing with the size of the enrollment. 
This starts to happen at the accuracy target 0.01\%.
At the same time, FNIR is close to zero in these cases.
Still, the divergence between search methods in terms of FPIR or FNIR is negligible in all cases, revealing how 1:First affects specifically the E-FPIR score.

A similar overview is shown in Figure \ref{fig:acc_ve_open_thr}, which contains VeriEye performance scores in open-set scenarios.
All the scores present the same trends found in IrisBee results. 
As expected from a high accuracy matcher, FNIR for VeriEye is below $10\%$ in all cases. 
On the other hand, its performance in the unenrolled portion of the probe set (FPIR) is worse than IrisBee by as much as $24\%$.
1:First search also had worse overall results than IrisBee regarding the E-FPIR, but not as accentuated as FPIR.

\subsection{Enrollment Permutations}

Contrary to what happens in 1:N search, 1:First accuracy can be affected by the ordering of the enrollment. 
To better understand how this effect could interfere in search accuracy, we perform experiments in which the probe set is presented to different permutations of the same enrollment.
These experiments are also performed on both matchers, using the same matching thresholds presented so far.

\begin{figure}[!t]
    \centering
    
    \begin{subfigure}[d]{\textwidth}
        \centering
        \includegraphics[width=\textwidth]{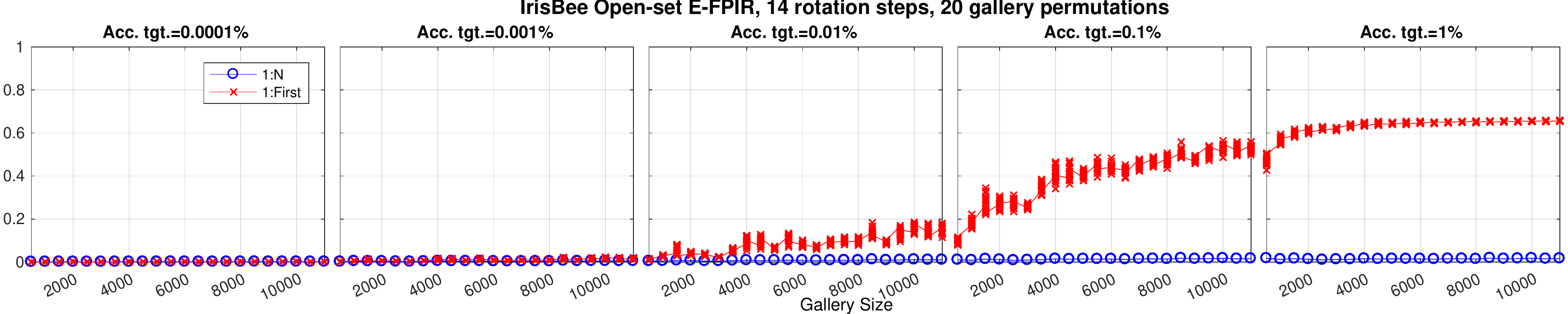}
        \caption{IrisBee}
        \label{fig:acc_ib_open_perm_thr}
    \end{subfigure}

    \begin{subfigure}[d]{\textwidth}
        \centering
        \includegraphics[width=\textwidth]{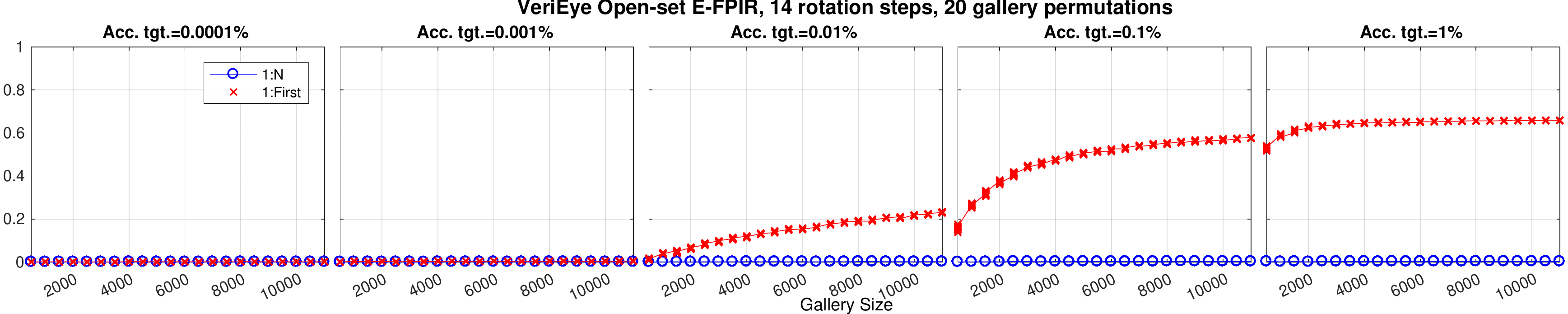}
        \caption{VeriEye}
        \label{fig:acc_ve_open_perm_thr}
    \end{subfigure}

    \caption{E-FPIR identification performance in open-set scenario, at different accuracy targets, with 20 random enrollment permutations.}
\end{figure}

Figure \ref{fig:acc_ib_open_perm_thr} presents performance scores for IrisBee in an open-set scenario across 20 random enrollment permutations. 
In this figure, the distribution of scores is plotted for each enrollment size.
As expected, 1:N results show no variance in performance. 
Using 1:First search however, some degree of variance in E-FPIR exists, if the system accuracy target is permissive enough (E-FPIR at accuracy target 0.01\% or less strict).
At the highest tolerance setting (E-FPIR at accuracy target 1\%), 1:First performance is so degraded that even variance is very small.

The general tendencies for both search methods however, remain the same: performance degradation starts to appear at a moderate level of strictness (0.01\%), and gets worse when the system is more lenient (0.1\% and 1\%).
1:First presents a moderate amount of variation in E-FPIR, and variance seems to decrease as the enrollment size increases.

In general, the same trends found in IrisBee results are seen in VeriEye. 
Comparing these results with IrisBee, the most distinguishing case is at target accuracy 0.01\%, where 1:First scores for VeriEye are in general higher than for IrisBee, while FNIR is under 5\% for both matchers.
In more tolerant levels (target accuracy 0.1\% and 1\%), the difference in 1:First error rates is not so accentuated, but is still a little higher in VeriEye than in IrisBee.

\begin{figure}[!tb]
    \centering
    \includegraphics[width=\textwidth]{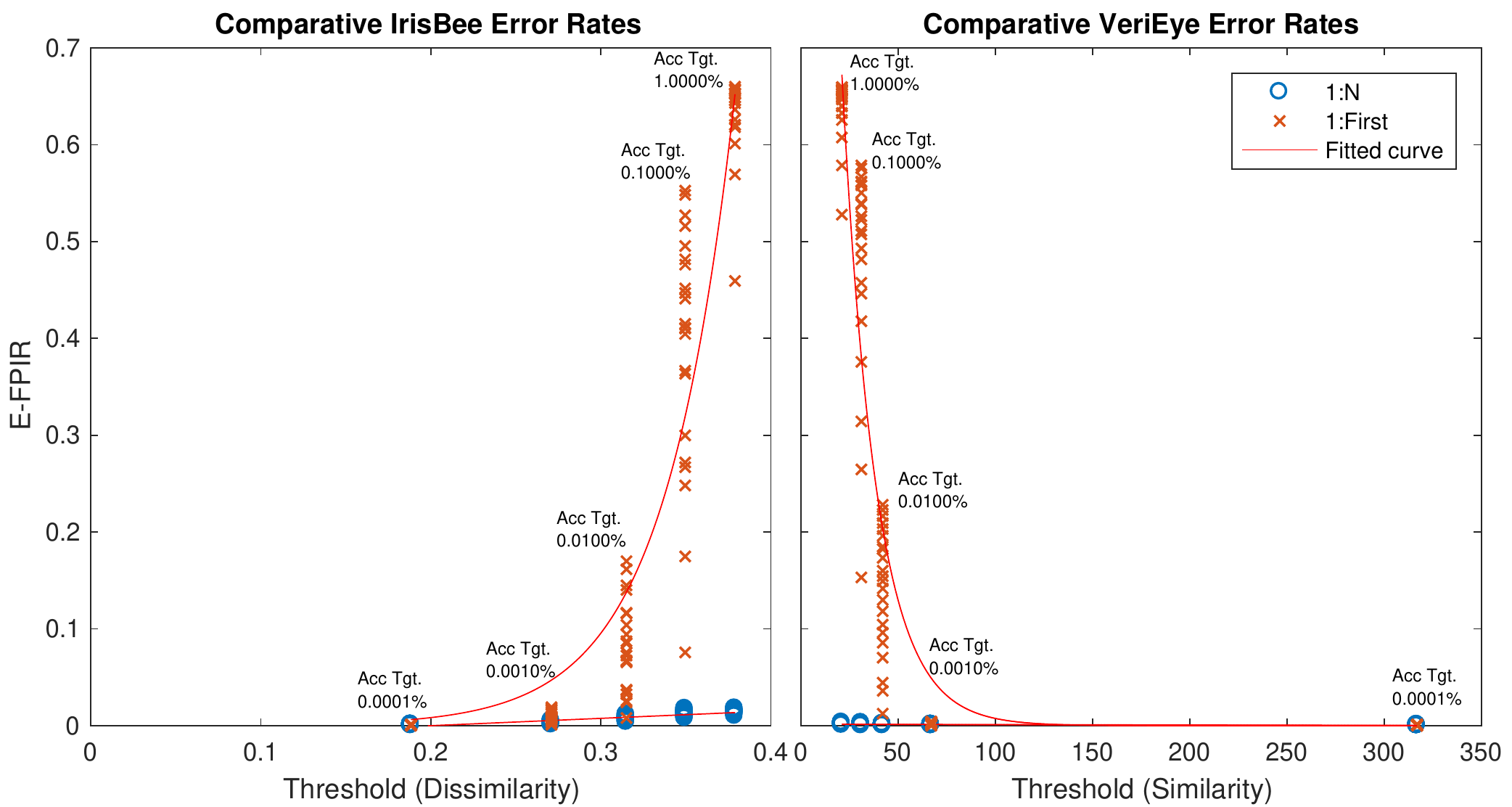}
    \caption{E-FPIR progression as the comparison threshold is adjusted: while using 1:N E-FPIR grows linearly, 1:First search results in an exponential growth when the threshold is made more lenient. }
    \label{fig:efpir_by_threshold}
\end{figure}

Additionally, we observe the dynamics of E-FPIR as the matching threshold is relaxed.
Figure \ref{fig:efpir_by_threshold} illustrates the result of this experiment. 
Note that the threshold scale ($x$ axis) of VeriEye is inverted in relation to IrisBee, because it uses a similarity score (as explained in Section \ref{sec:thressel}).
This graph shows how differently E-FPIR behaves under the different search methods.
Using 1:N search E-FPIR grows linearly as the threshold is made more lenient, and it never grows above 2\%.
On the other hand, when performing 1:First search, E-FPIR grows exponentially as we adjust the threshold to a more permissive setting.

On the other hand, the graph suggests it is possible to find an optimal threshold setting, where 1:First performance loss is minimal.
For instance, if the target accuracy of 0.001\% is selected, we can expect no serious degradation of E-FPIR, while still maintaining acceptable FNIR.

\subsection{Speed \emph{vs.} Accuracy}

The whole rationale behind 1:First search is that it is more efficient than 1:N. 
However, the usual notion is that 1:First is faster ``on average'', but no quantification is given.
Our experiments use the number of iris template comparisons as a speed measurement.
In order to be able to compare these measurements, we normalize the number of comparisons performed in the search by the total possible comparisons.
This way, 1:N will always present a normalized number of comparisons of 1 -- because the whole enrollment is scanned. 
On the other hand, 1:First will yield a score smaller than 1, since the search is interrupted before reaching the end of the enrollment.

\begin{figure}[!t]
    \centering
    \includegraphics[width=\textwidth]{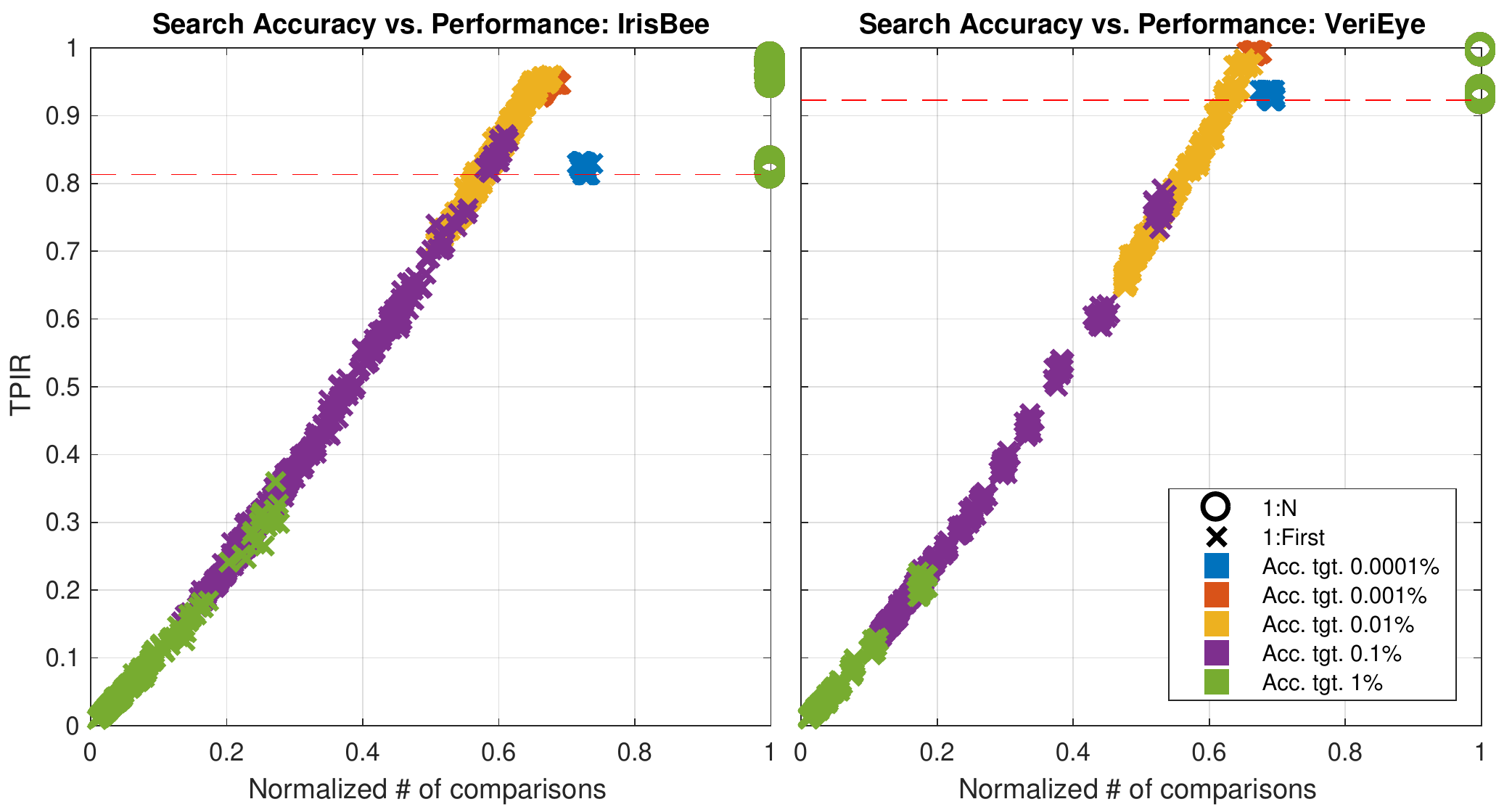}
    \caption{Relation between search speed and accuracy compared between search methods.}
    \label{fig:speed_eval}
\end{figure}

Figure \ref{fig:speed_eval} presents graphs that relate the search accuracy (in terms of TPIR), to search performance (in normalized number of comparisons).
All the 1:N results (circles) overlap almost perfectly in the top right corner of the plots.
For this reason it is not possible to see 1:N results for all accuracy targets.
At the same time, 1:First (Xs), scatter in a diagonal that ends near the origin. 
As the accuracy target is relaxed, a proportional reduction in accuracy and number of comparisons is observed.
Such degradation does not occur in the two strictest settings (blue and orange).

The red dotted line represents the worst accuracy obtained by 1:N search.
All 1:First results for accuracy targets above 0.001\% are above this line, meaning they produce accuracies comparable to 1:N, but performing only $\sim50-70\%$ of comparisons.
Also, at these accuracy targets, the linear degradation trend does not seem to take place.

\section{Conclusions}
\label{sec:conclusions}

The general tendency that can be apprehended in these experiments is that other parameters kept the same, 1:First search usually has worse accuracy than 1:N search.
1:First search accuracy degrades more quickly with increased enrollment size than does 1:N accuracy: this trend happened in all experiments and scenarios, specifically under the E-FPIR metric.

Making the system tolerance for identification more strict can lower the FPIR, but at the cost of increasing the FNIR.
If the system is set to a very restrictive accuracy target of 0.0001\%, there is no perceptible difference between the search methods in terms of FPIR, but FNIR can start becoming too high for practical use.

Our experiments with larger enrollment sizes revealed no contradicting trends from the experiments with smaller galleries \citep{kuehlkamp2016_1first}.
In fact, it illustrates how 1:First search accuracy degradation is closely connected to the enrollment size, while the same does not happen with 1:N.
Previous work shows how FPIR grows in small galleries \citep{kuehlkamp2016_1first}, and although at a lower rate than 1:First, 1:N accuracy also degrades.
Experiments with larger galleries showed that the accuracy degradation in 1:N is a phenomenon specific of small galleries: once false positive identification stabilizes at galleries of $\sim1500$ subjects, it does not increase again, regardless of the enrollment size (Figs. \ref{fig:acc_ib_closed_thr}, \ref{fig:acc_ve_closed_thr}, \ref{fig:acc_ib_open_thr} and \ref{fig:acc_ve_open_thr}).
An image quality assessment would certainly improve the understanding of 1:First performance in relation to 1:N, and could be an important topic for future research.

Perhaps one of the most intriguing results regards the behavior of matchers in open-set scenarios: although the error rate for the enrolled portion of the probe set remains similar to what was previously found, the unenrolled error rate revealed a different trend.
Unlike with the E-FPIR, FPIR calculation showed there is no difference in accuracy between 1:N and 1:First in the unenrolled portion of the probe set.
However not obvious, this fact is explained by the way 1:First works. 
If a 1:First search for a probe is interrupted, one of two situations occur: a) a positive identification is declared, if the biometric reference corresponds to the same subject as the biometric probe; or b) a false positive identification is declared, if the biometric reference and biometric probe do not correspond to the same subject.
Yet, when the subject corresponding to a biometric probe is unenrolled (has no corresponding biometric reference in a database), it does not matter if the search is interrupted, because the only possible outcomes are either a true negative or a false positive identification.

The last set of experiments confirmed all the previous tendencies for the behavior of 1:N and 1:First searches. Performing 1:First search against different permutations of the same enrollment introduces some degree of variance into the results, but the standard error from the mean does not exceed 6\%.

Experiments show that E-FPIR performance degradation in 1:First search has an exponential relation with the threshold relaxation. 
It is possible, though, to find an optimal comparison threshold at a reasonable accuracy setting and this way be able to use 1:First search with little performance degradation in comparison to 1:N.
In our experiments, this occurred for accuracy targets of 0.001\% or 0.0001\%. 
Nevertheless, it is important to keep in mind that at these accuracy targets, FNIR starts to increase and may impair system usability.

Finally, our experiments show that 1:First search is able to achieve accuracy comparable to 1:N while performing only $\sim50-70\%$ of the comparisons.
However, if accuracy target is relaxed, performance degrades proportionally to the reduction in search comparisons.

\section{Acknowledgements}
\label{sec:ack}
The authors thank the valuable contributions given by Dr. Adam Czajka.
This research was partially supported by the Brazilian Ministry of Education -- CAPES through process BEX 12976/13-0. 

\bibliographystyle{elsarticle-num-names}

\bibliography{sample}

\begin{thebibliography}{21}
\providecommand{\natexlab}[1]{#1}
\providecommand{\url}[1]{\texttt{#1}}
\providecommand{\urlprefix}{URL }
\expandafter\ifx\csname urlstyle\endcsname\relax
  \providecommand{\doi}[1]{doi:\discretionary{}{}{}#1}\else
  \providecommand{\doi}[1]{doi:\discretionary{}{}{}\begingroup
  \urlstyle{rm}\url{#1}\endgroup}\fi
\providecommand{\bibinfo}[2]{#2}

\bibitem[{Jain et~al.(2011)Jain, Ross, and Nandakumar}]{Jain2011}
\bibinfo{author}{A.~K. Jain}, \bibinfo{author}{A.~A. Ross},
  \bibinfo{author}{K.~Nandakumar}, \bibinfo{title}{{Introduction to
  Biometrics}}, \bibinfo{publisher}{Springer}, ISBN
  \bibinfo{isbn}{9780387773254}, \bibinfo{year}{2011}.

\bibitem[{Daugman(2008)}]{Daugman2008}
\bibinfo{author}{J.~Daugman}, \bibinfo{title}{Iris Recognition}, in:
  \bibinfo{editor}{A.~Jain}, \bibinfo{editor}{P.~Flynn},
  \bibinfo{editor}{A.~Ross} (Eds.), \bibinfo{booktitle}{Handbook of
  Biometrics}, chap.~\bibinfo{chapter}{4}, \bibinfo{publisher}{Springer},
  \bibinfo{year}{2008}.

\bibitem[{Bengali(2017)}]{latimes2017}
\bibinfo{author}{S.~Bengali}, \bibinfo{title}{India is building a biometric
  database for 1.3 billion people -- and enrollment is mandatory},
  \bibinfo{howpublished}{\textit{Los Angeles Times}, 11 May 2017. Available
  \url{http://www.latimes.com/world/la-fg-india-database-2017-story.html} [Last
  accessed in: 14 May 2017.]}, \bibinfo{year}{2017}.

\bibitem[{Interpeace(2016)}]{interpeace2016somaliland}
\bibinfo{author}{Interpeace}, \bibinfo{title}{Somaliland successfully launches
  voter registration}, \bibinfo{howpublished}{\textit{Interpeace}, 21 January
  2016. Available
  \url{http://www.interpeace.org/2016/01/somaliland-successfully-launches-voter-registration/}
  [Last accessed in: 06 June 2017.]}, \bibinfo{year}{2016}.

\bibitem[{Sandhana(2014)}]{newsci2014somaliland}
\bibinfo{author}{L.~Sandhana}, \bibinfo{title}{Iris register to eyedentify
  voting fraud in Somaliland}, \bibinfo{howpublished}{\textit{New Scientist},
  03 September 2014. Available
  \url{{https://www.newscientist.com/article/mg22329854-400-iris-register-to-eyedenti
  fy-voting-fraud-in-somaliland/}} [Last accessed in: 06 June 2017.]},
  \bibinfo{year}{2014}.

\bibitem[{Bowyer et~al.(2015)Bowyer, Ortiz, and Sgroi}]{bowyer2015somaliland}
\bibinfo{author}{K.~W. Bowyer}, \bibinfo{author}{E.~Ortiz},
  \bibinfo{author}{A.~Sgroi}, \bibinfo{title}{Trial Somaliland voting register
  de-duplication using iris recognition}, in: \bibinfo{booktitle}{2015 11th
  IEEE International Conference and Workshops on Automatic Face and Gesture
  Recognition (FG)}, vol.~\bibinfo{volume}{02}, \bibinfo{pages}{1--8},
  \doi{\bibinfo{doi}{10.1109/FG.2015.7284833}}, \bibinfo{year}{2015}.

\bibitem[{Daugman(2015)}]{Daugman2015airports}
\bibinfo{author}{J.~Daugman}, \bibinfo{title}{Iris Recognition at Airports and
  Border Crossings}, in: \bibinfo{editor}{S.~Z. Li}, \bibinfo{editor}{A.~K.
  Jain} (Eds.), \bibinfo{booktitle}{Encyclopedia of Biometrics},
  \bibinfo{publisher}{Springer US}, \bibinfo{address}{Boston, MA}, ISBN
  \bibinfo{isbn}{978-1-4899-7488-4}, \bibinfo{pages}{998--1004},
  \doi{\bibinfo{doi}{10.1007/978-1-4899-7488-4\_24}},
  \urlprefix\url{http://dx.doi.org/10.1007/978-1-4899-7488-4\_24},
  \bibinfo{year}{2015}.

\bibitem[{{Canada Border Services Agency}(2017)}]{nexus1}
\bibinfo{author}{{Canada Border Services Agency}}, \bibinfo{title}{{Travel
  often? Apply for NEXUS}}, \bibinfo{howpublished}{{Canada Border Services
  Agency} website. Available at
  \url{https://www.cbsa-asfc.gc.ca/prog/nexus/menu-eng.html} [Last accessed in:
  13 January 2018.]}, \bibinfo{year}{2017}.

\bibitem[{Kuehlkamp and Bowyer(2016)}]{kuehlkamp2016_1first}
\bibinfo{author}{A.~Kuehlkamp}, \bibinfo{author}{K.~W. Bowyer},
  \bibinfo{title}{An analysis of 1-to-first matching in iris recognition}, in:
  \bibinfo{booktitle}{2016 IEEE Winter Conference on Applications of Computer
  Vision (WACV)}, \doi{\bibinfo{doi}{10.1109/WACV.2016.7477687}},
  \bibinfo{year}{2016}.

\bibitem[{{National Science \& Technology Council, Subcommittee on
  Biometrics}(2006)}]{Biometrics2006}
\bibinfo{author}{{National Science \& Technology Council, Subcommittee on
  Biometrics}}, \bibinfo{title}{{Biometrics Glossary}},
  \bibinfo{howpublished}{Available at
  \url{http://www.nws-sa.com/biometrics/biooverview.pdf} [Last accessed in: 07
  July 2017.]}, \bibinfo{year}{2006}.

\bibitem[{ISO 19795-1:2006(2006)}]{ISO19795}
ISO 19795-1:2006, \bibinfo{title}{{International Standard ISO/IEC 19795-1
  Biometric performance testing and reporting — Part 1}},
  \bibinfo{type}{Standard}, \bibinfo{institution}{International Organization
  for Standardization}, \bibinfo{address}{Geneva, CH}, \bibinfo{year}{2006}.

\bibitem[{Mukherjee and Ross(2008)}]{Mukherjee2008}
\bibinfo{author}{R.~Mukherjee}, \bibinfo{author}{A.~Ross},
  \bibinfo{title}{{Indexing iris images}}, \bibinfo{journal}{19th International
  Conference on Pattern Recognition} ~(\bibinfo{number}{December}), ISSN
  \bibinfo{issn}{1051-4651}, \doi{\bibinfo{doi}{10.1109/ICPR.2008.4761880}}.

\bibitem[{Proen{\c{c}}a(2013)}]{Proenca2013}
\bibinfo{author}{H.~Proen{\c{c}}a}, \bibinfo{title}{{Iris biometrics: Indexing
  and retrieving heavily degraded data}}, \bibinfo{journal}{IEEE Transactions
  on Information Forensics and Security}
  \bibinfo{volume}{8}~(\bibinfo{number}{12}) (\bibinfo{year}{2013})
  \bibinfo{pages}{1975--1985}, ISSN \bibinfo{issn}{15566013},
  \doi{\bibinfo{doi}{10.1109/TIFS.2013.2283458}}.

\bibitem[{Rathgeb and Uhl(2010)}]{Rathgeb2010}
\bibinfo{author}{C.~Rathgeb}, \bibinfo{author}{A.~Uhl},
  \bibinfo{title}{Iris-Biometric Hash Generation for Biometric Database
  Indexing}, in: \bibinfo{booktitle}{2010 20th International Conference on
  Pattern Recognition}, ISSN \bibinfo{issn}{1051-4651},
  \bibinfo{pages}{2848--2851}, \doi{\bibinfo{doi}{10.1109/ICPR.2010.698}},
  \bibinfo{year}{2010}.

\bibitem[{Rakvic et~al.(2009)Rakvic, Ulis, Broussard, Ives, and
  Steiner}]{rakvic2009}
\bibinfo{author}{R.~N. Rakvic}, \bibinfo{author}{B.~J. Ulis},
  \bibinfo{author}{R.~P. Broussard}, \bibinfo{author}{R.~W. Ives},
  \bibinfo{author}{N.~Steiner}, \bibinfo{title}{Parallelizing iris
  recognition}, \bibinfo{journal}{IEEE Transactions on Information Forensics
  and Security} \bibinfo{volume}{4}~(\bibinfo{number}{4})
  (\bibinfo{year}{2009}) \bibinfo{pages}{812--823}.

\bibitem[{Hao et~al.(2008)Hao, Daugman, and Zieli{\'n}ski}]{hao2008}
\bibinfo{author}{F.~Hao}, \bibinfo{author}{J.~Daugman},
  \bibinfo{author}{P.~Zieli{\'n}ski}, \bibinfo{title}{A fast search algorithm
  for a large fuzzy database}, \bibinfo{journal}{IEEE Transactions on
  Information Forensics and Security} \bibinfo{volume}{3}~(\bibinfo{number}{2})
  (\bibinfo{year}{2008}) \bibinfo{pages}{203--212}.

\bibitem[{Ortiz and Bowyer(2015)}]{Ortiz2015}
\bibinfo{author}{E.~Ortiz}, \bibinfo{author}{K.~W. Bowyer},
  \bibinfo{title}{Exploratory analysis of an operational iris recognition
  dataset from a CBSA border-crossing application}, in:
  \bibinfo{booktitle}{2015 IEEE Conference on Computer Vision and Pattern
  Recognition Workshops (CVPRW)}, ISSN \bibinfo{issn}{2160-7508},
  \bibinfo{pages}{34--41}, \doi{\bibinfo{doi}{10.1109/CVPRW.2015.7301317}},
  \bibinfo{year}{2015}.

\bibitem[{Bowyer et~al.(2013)Bowyer, Hollingsworth, and
  Flynn}]{Bowyer2013survey}
\bibinfo{author}{K.~W. Bowyer}, \bibinfo{author}{K.~P. Hollingsworth},
  \bibinfo{author}{P.~J. Flynn}, \bibinfo{title}{A Survey of Iris Biometrics
  Research: 2008--2010}, in: \bibinfo{editor}{M.~J. Burge},
  \bibinfo{editor}{K.~W. Bowyer} (Eds.), \bibinfo{booktitle}{Handbook of Iris
  Recognition}, \bibinfo{publisher}{Springer London},
  \bibinfo{address}{London}, ISBN \bibinfo{isbn}{978-1-4471-4402-1},
  \bibinfo{pages}{15--54}, \doi{\bibinfo{doi}{{10.1007/978-1-4471-4402-1}}},
  \bibinfo{year}{2013}.

\bibitem[{Daugman(2006)}]{Daugman2006}
\bibinfo{author}{J.~Daugman}, \bibinfo{title}{{Probing the Uniqueness and
  Randomness of IrisCodes: Results From 200 Billion Iris Pair Comparisons}},
  \bibinfo{journal}{Proceedings of the IEEE}
  \bibinfo{volume}{94}~(\bibinfo{number}{11}) (\bibinfo{year}{2006})
  \bibinfo{pages}{1927--1935}, ISSN \bibinfo{issn}{0018-9219},
  \doi{\bibinfo{doi}{10.1109/JPROC.2006.884092}},
  \urlprefix\url{http://ieeexplore.ieee.org/lpdocs/epic03/wrapper.htm?arnumber=4052470}.

\bibitem[{Chumakov(2015)}]{Chumakov1}
\bibinfo{author}{M.~Chumakov}, \bibinfo{title}{{Canada Border Services
  Agency}}, \bibinfo{howpublished}{Private communication},
  \bibinfo{year}{2015}.

\bibitem[{Czajka et~al.(2017)Czajka, Bowyer, Krumdick, and {Vidal
  Mata}}]{Czajka2017_LRUD}
\bibinfo{author}{A.~Czajka}, \bibinfo{author}{K.~Bowyer},
  \bibinfo{author}{M.~Krumdick}, \bibinfo{author}{R.~{Vidal Mata}},
  \bibinfo{title}{{Recognition of image-orientation-based iris spoofing}},
  \bibinfo{journal}{IEEE Transactions on Information Forensics and Security}
  \bibinfo{volume}{6013}~(\bibinfo{number}{c}), ISSN \bibinfo{issn}{1556-6013},
  \doi{\bibinfo{doi}{10.1109/TIFS.2017.2701332}},
  \urlprefix\url{http://ieeexplore.ieee.org/document/7919203/}.

\end{thebibliography}

\end{document}